%% file: main.tex
\documentclass[final]{article}
\usepackage{graphicx} 
\usepackage[preprint,nonatbib]{compsust_2023}
\usepackage{subcaption}
\usepackage{mwe}

\usepackage{algpseudocode}
\usepackage{algorithm}

\usepackage{amsmath}
\usepackage{wrapfig}
\usepackage{comment}
\usepackage{booktabs}

\title{Aggregate Representation Measure for \\Predictive Model Reusability}

\author{
    Vishwesh Sangarya \\
    North Carolina State University \\
    \And
    Richard Bradford \\
    Collins Aerospace \\
    \And
    Jung-Eun Kim\thanks{Correspondence.} \\
    North Carolina State University 
}

\begin{document}

\maketitle

\begin{abstract}

In this paper, we propose a predictive quantifier to estimate the retraining cost of a trained model in distribution shifts. The proposed Aggregated Representation Measure (ARM) quantifies the change in the model's representation from the old to new data distribution. It provides, before actually retraining the model, a single concise index of resources - epochs, energy, and carbon emissions - required for the retraining. This enables reuse of a model with a much lower cost than training a new model from scratch. The experimental results indicate that ARM reasonably predicts retraining costs for varying noise intensities and enables comparisons among multiple model architectures to determine the most cost-effective and sustainable option.


\end{abstract}




\section{Introduction}
As deep neural networks are becoming increasingly prevalent in everyday applications and deployments, involving ever larger datasets, the compute requirements keep increasing with larger models, and their energy consumption becomes an issue. Recent research \cite{xu2023energy, strubell2019energy, xu2021survey, GARCIAMARTIN201975, yang2017designing, gholami2022survey} have addressed the energy efficiency of diverse neural network methods, as well as the issue of carbon emissions \cite{schmidt2021codecarbon, anthony2020carbontracker, lacoste2019quantifying}. At the same time, there is an ongoing need for deployed neural networks to respond to changes in their environment. When a deep learning model sees distributional shifts in data, it is desirable to adapt itself to the change. To develop models robust to such distributional shifts, existing solutions \cite{djolonga2021robustness, andreassen2021evolution} train the model from scratch with larger models, larger training data, and longer training duration. However, in such approaches, the resource requirements are costly.


One of the sustainable solutions to this problem is \emph{reusing} an existing model to adapt to the new environment. By retraining previously-trained models on the new distributions, energy consumption and carbon emissions will be significantly reduced. Then there are potential questions to consider: (i) How do different models adapt to a certain distributional shift? (ii) How does a given model adapt to different levels of noise or corruption? (iii) Is it possible to predict the behavior of a model before expending the cost of retraining and adapting it?

In order to quantitatively answer those questions, we propose a predictive reusability quantifier, Aggregate Representation Measure (ARM). ARM works by quantifying the change in a model's representation for new distributional shifts. In particular, ARM quantifies the change in representation for each layer and then aggregates it for the entire model. It provides a single concise value that can predict the retraining efforts required to adapt a model to a distributional shift. Using ARM, the energy consumption and carbon emission can be predicted before expending the retraining costs. We show that ARM requires only one forward pass through the model, and we provide evidence of how it strongly correlates to retraining measures, training epochs, energy, and carbon emissions. ARM not only helps predict the behavior of a model for different levels of noise but also allows comparisons among different models, thus enabling better decisions on the model type to be deployed.

\section{Related work}
Since \cite{hendrycks2019benchmarking}, several techniques, architectures and training methodologies have emerged to improve model robustness. While certain models architectures and training methods do generate robust models, there is a tradeoff with regard to model size or the size of training data used as shown by \cite{djolonga2021robustness, andreassen2021evolution}. The use of large models for real-world deployments and the hyperparameter search involved with training these large models, with the increased dataset size is a resource-exhaustive process. Several research works \cite{geirhos2020generalisation, yin2020fourier, ford2019adversarial} have shown that there is a non-uniform improvement in robustness to the different distribution shifts, in some cases improvement on one type of noise or corruption results in decreased performance on a distributional shift. In general, even with uneven, non-uniform gains on certain distributions, with decreased improvements on other distributions, the training and augmentation techniques \cite{drenkow2022systematic, hendrycks2020augmix, liu2022randommix, zhang2018mixup, kimICML20, lee2020smoothmix} are computationally heavy and require training a model from scratch.
Methods which using test time adaptation \cite{lim2023ttn, niu2022efficient, goyal2022testtime, wang2022continual} exhibit only marginal improvements in model robustness and fail to provide substantial benefits in scenarios with elevated noise levels. If the test time information is insufficient for adapting the model's prediction, these methods fail to provide accurate and confident outputs during inference.

With the pressing need for sustainable development of deep learning networks, several works have focused on monitoring the energy and carbon emissions of neural network training. Several works \cite{schmidt2021codecarbon, lacoste2019quantifying, GARCIAMARTIN201975, anthony2020carbontracker} focus on neural network energy consumption and carbon emissions.
\cite{strubell2019energy, xu2021survey, xu2023energy} highlight the need for measuring energy and carbon emissions of neural network training and deployment. 
They call attention to the significantly high energy usage and carbon emissions that are a part of neural network training and hyper parameter search mechanisms involved.
Works such as \cite{stacke2020measuring} use the change in layer representation to study pathology data and focus their work to individual layers of a model to show it correlates to accuracy loss on domain shifts.


\section{Aggregate representation measure}
  \begin{wrapfigure}{R}{0.5\textwidth}
    \begin{minipage}{0.5\textwidth}
\vspace{-0.7cm}
\begin{algorithm}[H]
\caption{Calculate $f_{l,k}$}
\begin{algorithmic}[1]
\State $F_{l,k} \gets \{\}$
\For{$x \gets 1$ to $n$}  
    \State $avg\_activation \gets \frac{1}{h \cdot w} \sum_{i=1,j=1}^{h,w} I_{x_{i,j}}$
    \State $F_{l,k} \gets F_{l,k} \cup avg\_activation$  
\EndFor
\State $f_{l,k} \gets PF(F_{l,k})$
\end{algorithmic}
\label{alg:algorighm}
\end{algorithm}
\vspace{-0.6cm}
    \end{minipage}
  \end{wrapfigure}
Aggregated representation measure (ARM) makes use of the model's change in representation between the data it was trained on and the new distribution. ARM is calculated per layer of the model and then averaged to give a single scalar value that describes how much the shift is in the model's representation. To be able to capture the representation for each data, we perform one forward pass of the entire dataset through the model. During the forward pass, the activation outputs of each filter or neuron are collected. For convolutional layers, to reduce the memory requirements, the activation output is averaged. For each layer $l$ with $k$ filters, the probability function of a given filter $f_{l,k}$ is obtained by iterating over the dataset of size $n$. We introduce the filter probability distribution, $f_{l,k}$ in Algorithm~\ref{alg:algorighm}, where $F_{l,k}$ is a set containing the averaged activation values for the dataset. $PF$ represents the probability distribution function computed on the complete set of activation data. For each filter/neuron in a given layer, the activation outputs for the entire dataset are collected, as shown in Algorithm~\ref{alg:algorighm}. For each data sample $I_x$, the activation for the sample is summed and averaged, where $h$ and $w$ represent the height and width of the activation.

For each layer $l$, the layer probability $P_l$ is obtained by:
\begin{equation}
    P_{l} = \frac{1}{n_l} \sum_{k=1}^{n_l} f_{l,k}
    \label{eq:layer_prob}
\end{equation}
where $n_l$ represents the number of filters/neurons in a given layer $l$. For each layer, the filter/neuron probability distributions are summed and averaged to produce a layer probability distribution as shown in~\eqref{eq:layer_prob}, where $P_l$ is the averaged layer probability distribution for layer $l$. This operation is performed for the second distribution shifted dataset. 

The Aggregate Representation Measure, ARM, is obtained as follows:
\begin{equation}
    ARM = \frac{1}{L} \sum_{l=1}^{L} {WD(P_{l, d1}, P_{l, d2}})
    \label{eq:final_measure}
\end{equation} 

$WD$ represents Wasserstein distance between the two probability distributions. The final result - ARM is calculated by finding the average of the Wasserstein distance between each layer's corresponding probability distribution for the entire model. As the layer probability distribution may have small differences, the Wasserstein metric, with its displacement-based measurement, offers a fine-grained and more sensitive measure, making it preferable over other measures such as Jensen-Shannon divergence for capturing subtle distinctions. \eqref{eq:final_measure} provides the final aggregated measure for a given model with $L$ layers, where $P_{l, d1}$ and $P_{l, d2}$ represent the probability distributions for a given layer $l$ over the original data $d1$ and new data $d2$ by distributional shift, respectively.

\section{Retraining measures}
Retraining measures quantify the effort and resources required to adapt a model to a new distribution. The retraining measures used are epochs, energy consumption, and carbon emissions which help quantify the sustainability of adapting a model. To measure the number of epochs required for retraining a model, we set a minimum required accuracy for each dataset and train the model until it reaches the accuracy level. To measure carbon emission and energy consumption, we use \cite{schmidt2021codecarbon}, which makes use of the energy consumed for the retraining and the location of the energy generated to calculate the likely weight of carbon compounds emitted into the atmosphere. For a coarse-grained analysis, ARM is capable of predicting the overall global gradient norm of a model during retraining. \cite{agarwal2022estimating} shows that gradients represent the difficulty of samples. This is useful, as the gradient norm as a retraining measure helps understand the depth and path of the loss landscape for each model, with a lower overall gradient norm resulting in faster convergence, as the model is able to adapt to the new distribution faster. A lower overall gradient norm also represents that the model's current parameter space is closer to the new parameter space post retraining. We reserve gradient norm and standard learning rate experiment analysis for future work.

\section{Experiments and Results}
In the experiments, three datasets are used, CIFAR10, CIFAR100, and SVHN, with 3 noise types - Image Blur, Gaussian noise, and Salt-Pepper noise. For each noise type, 7-9 different noise levels are employed. In the charts, Noise 1 represents the lowest noise level (intensity). The highest level of noise is comparable to severity level 4 in \cite{hendrycks2019benchmarking}. The large number of noise intervals is to provide detailed working evidence of ARM and its correlation to the retraining measures. We explore different model architectures, ResNet, VGG, GoogLeNet, and MobileNetV2. 
To reuse an existing model, we train a randomly initialized model on the original data distribution until it reaches the required accuracy for each dataset. All experiment results are an average of three runs.

\begin{figure}[t]
    \centering
    \begin{subfigure}{0.47\textwidth}
        \centering
        \includegraphics[width=\textwidth]{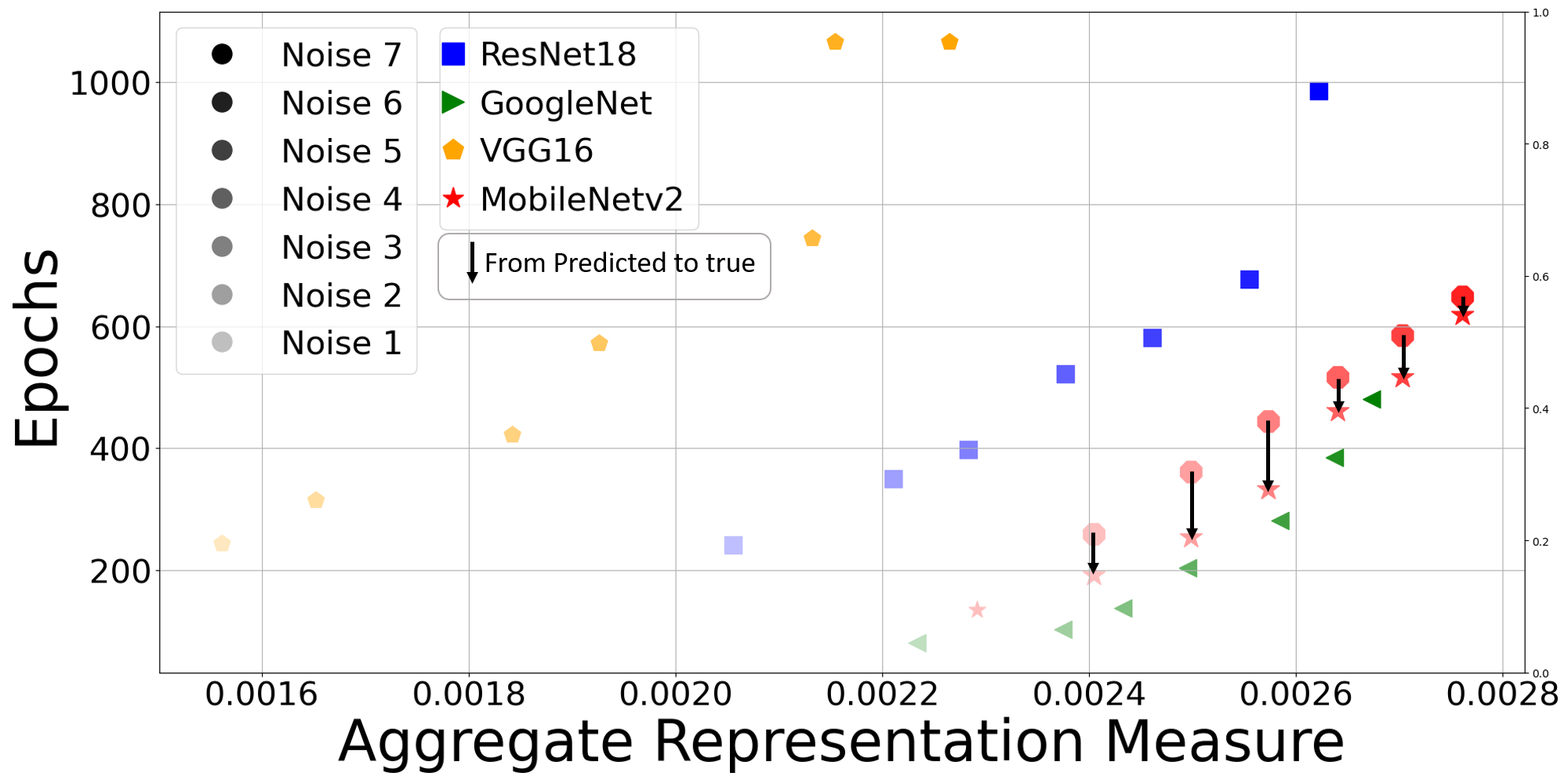}
        \caption{Predicting MobileNet for 6 noise levels}
    \end{subfigure}
    \begin{subfigure}{0.47\textwidth}
        \centering
        \includegraphics[width=\textwidth]{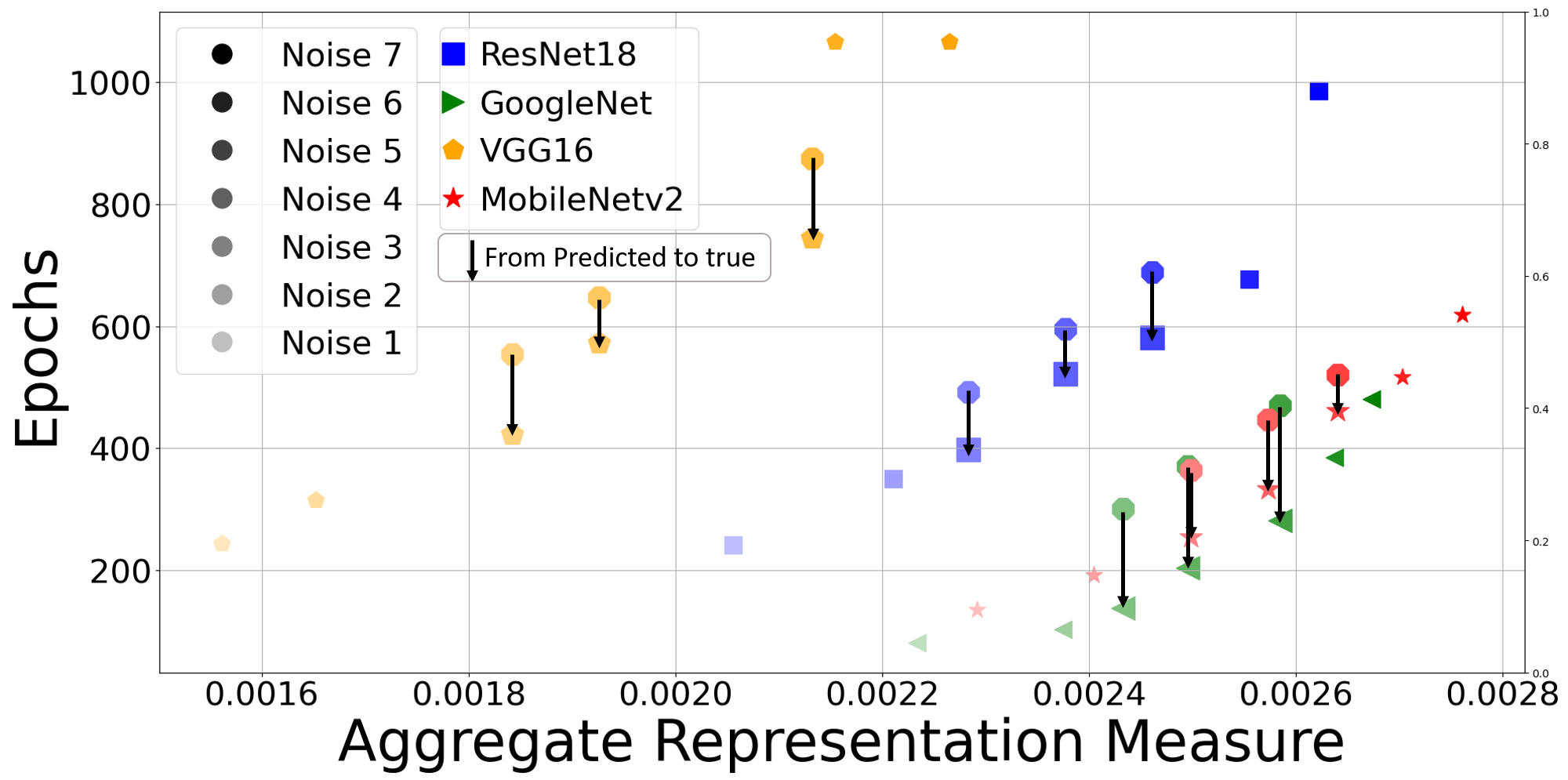}
        \caption{Predicting all models for 3 noise levels}
    \end{subfigure}
    \caption{Representation measure as a predictive metric}
    \label{fig:MNV_prediction}
\end{figure}

\begin{table}[htbp]
\footnotesize
        \centering
        \begin{tabular}{ccccccc}
             \toprule
             Model & Coefficient & p-value & Coefficient & p-value & Coefficient & p-value \\
             \midrule
             \multicolumn{1}{c}{ } &\multicolumn{2}{c}{Salt-Pepper} & \multicolumn{2}{c}{Gaussian} & \multicolumn{2}{c}{Blur} \\
             \midrule
             GoogLeNet & 0.92 & 0.00041 & 0.93 & 0.002 & 0.85 & 0.015 \\
             \hline
             ResNet18 & 0.93 & 0.00024 & 0.94 & 0.0014 & 0.92 & 0.0030 \\
             \hline
             MobileNetV2 & 0.91 & 0.00061 & 0.97 & 0.00016 & 0.87 & 0.0091 \\
             \hline
             VGG16 & 0.94 & 0.0014 &  0.95 & 0.0008 & 0.90 & 0.0051 \\
             \bottomrule
        \end{tabular}
        \vspace{0.1cm}
        \caption{Pearson correlation between epochs and ARM - CIFAR10 Dataset}
        \label{tab:correlation_table_C10_main}
\end{table}

Since various models are susceptible to different ranges and learning rate schedules, finding the most optimal learning rate plan will require computationally expensive hyperparameter search and tuning. The retrieved learning rate schedule may not be the most optimal one as it is difficult to verify how the learning rate needs to be adapted to the different loss and gradient landscapes for each model. To provide a fine-grained analysis regarding the number of epochs a model requires to adapt to a new data distribution, we set the learning rate to be extremely small for all models and experiments, in the order of $1e-4$ to $1e-6$. 

We perform experiments on Gaussian noise using ResNet18, GoogLeNet and VGG16 to collect the data, which exhibits the linear relation between ARM and epochs. We conduct a prediction on a new model, MobileNetv2, of which data is not used to model a regression predictor. We retrain MobileNetv2 on the first level of noise and obtain a starting value to model a regression. This starting value is used with the regression predictor to predict the epochs needed by MobileNetV2 to adapt to the remaining six levels of noise. The actual and predicted retraining epochs are presented in Fig.~\ref{fig:MNV_prediction} (a) which shows that the predicted values are a fair prediction of a new model's epochs required for retraining and the likely behavior to different noise levels.


We perform an experiment to predict additional and unseen noise levels for each of the 4 models. We use the data collected for the initial two and final two Gaussian noise levels for each model when modeling the regression predictor. Using this predictor, we predict the retraining epochs of each model for the three intermediate noise levels - noise levels 3, 4, and 5. Fig.~\ref{fig:MNV_prediction} (b) shows the true and predicted epochs for each model. In all our experiments, the ARM model consistently over-estimates the retraining cost; we see this as a feature, in that the estimates are conservative rather than falsely optimistic.  The predicted values are fairly close to the actual values considering that a single regression predictor was used for all models. With pre-existing or additional data of model's retraining and ARM values, highly accurate predictors for each individual model can be devised. Another point to pay attention to in Fig.~\ref{fig:MNV_prediction} is the inter-model comparison. For the same level of noise, GoogLeNet and MobileNetV2 adapt much faster as compared to ResNet18, and ResNet18 adapts faster than VGG16. This comparison helps predict and compare the adaptability and reusability among different model architectures and trained models.

Table~\ref{tab:correlation_table_C10_main} shows the Pearson correlation coefficient between ARM and epochs. The high Pearson correlation coefficient and low p-value indicate a strong positive relation between ARM and retraining epochs, suggesting that ARM is a suitable predictive measure to determine retraining costs.

\begin{figure*}
    \centering
    \includegraphics[width=0.9\textwidth]{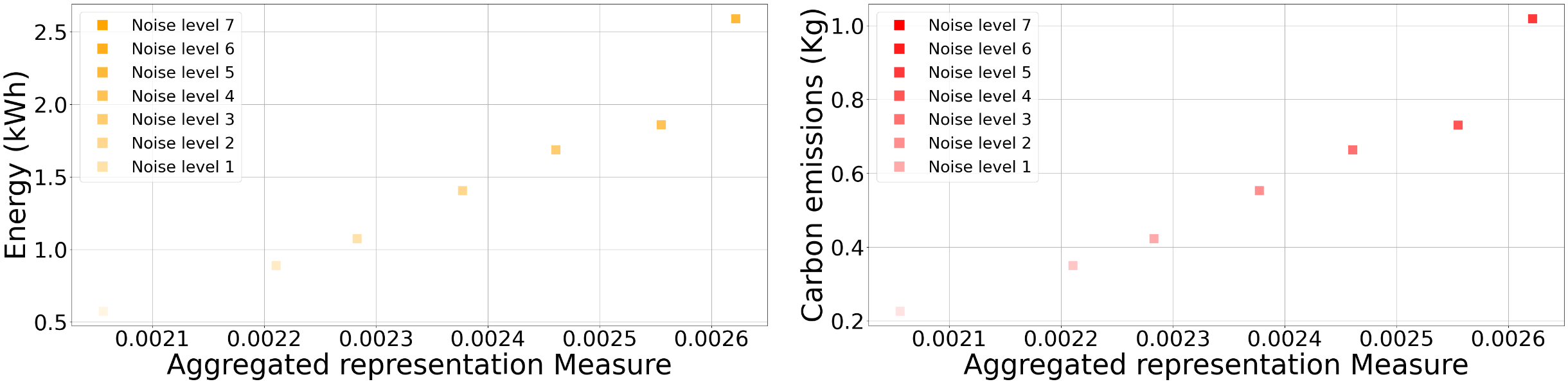}
    \caption{Energy and Carbon emissions of retraining ResNet18 to different levels of Gaussian noise}%
    \label{fig:EnergyCO2}
\end{figure*}

With the main objective being sustainable re-usability, we measure the energy and carbon emissions which have a strong positive correlation with the measure. Fig.~\ref{fig:EnergyCO2} depicts the linear increasing trend of ARM with retraining energy usage and carbon emissions for ResNet18.

Additional experiment results on CIFAR10, CIFAR100, and SVHN are provided in the Appendix. 

\section{Conclusion}
We have presented a novel metric to predict the cost of retraining a model to new distributional shifts. Our proposed measure, a predictive quantifier of the reusability of trained models, will help users make informed decisions. We demonstrated the correlation and predictive ability of ARM with epochs, energy, and carbon emissions, which indicates the effectiveness of ARM to predict a model's behavior. ARM enables intra-model comparison to different noise levels and inter-model comparison to select the most adaptable and sustainable model.

\acksection
This publication is based upon work supported by the National Science Foundation under Grant No. 1945541 (transferred and extended to No. 2302610). Any opinions, findings, and conclusions or recommendations expressed in this material are those of the author(s) and do not necessarily reflect the views of the National Science Foundation.

\bibliographystyle{plain}
\bibliography{cite}

\newpage
\appendix
\input{appendix}

\end{document}

%% file: appendix.tex
\section{Appendix: All experiment details - CIFAR10, CIFAR100, SVHN}

In this Appendix we provide the Pearson correlation coefficient, associated p-value and ARM vs. retraining epochs graphs for all models on CIFAR10, CIFAR100 and SVHN. 


\subsection{CIFAR10 Dataset}
\begin{figure}[ht]
    \centering
    \begin{subfigure}{0.45\textwidth}
        \includegraphics[width=\linewidth]{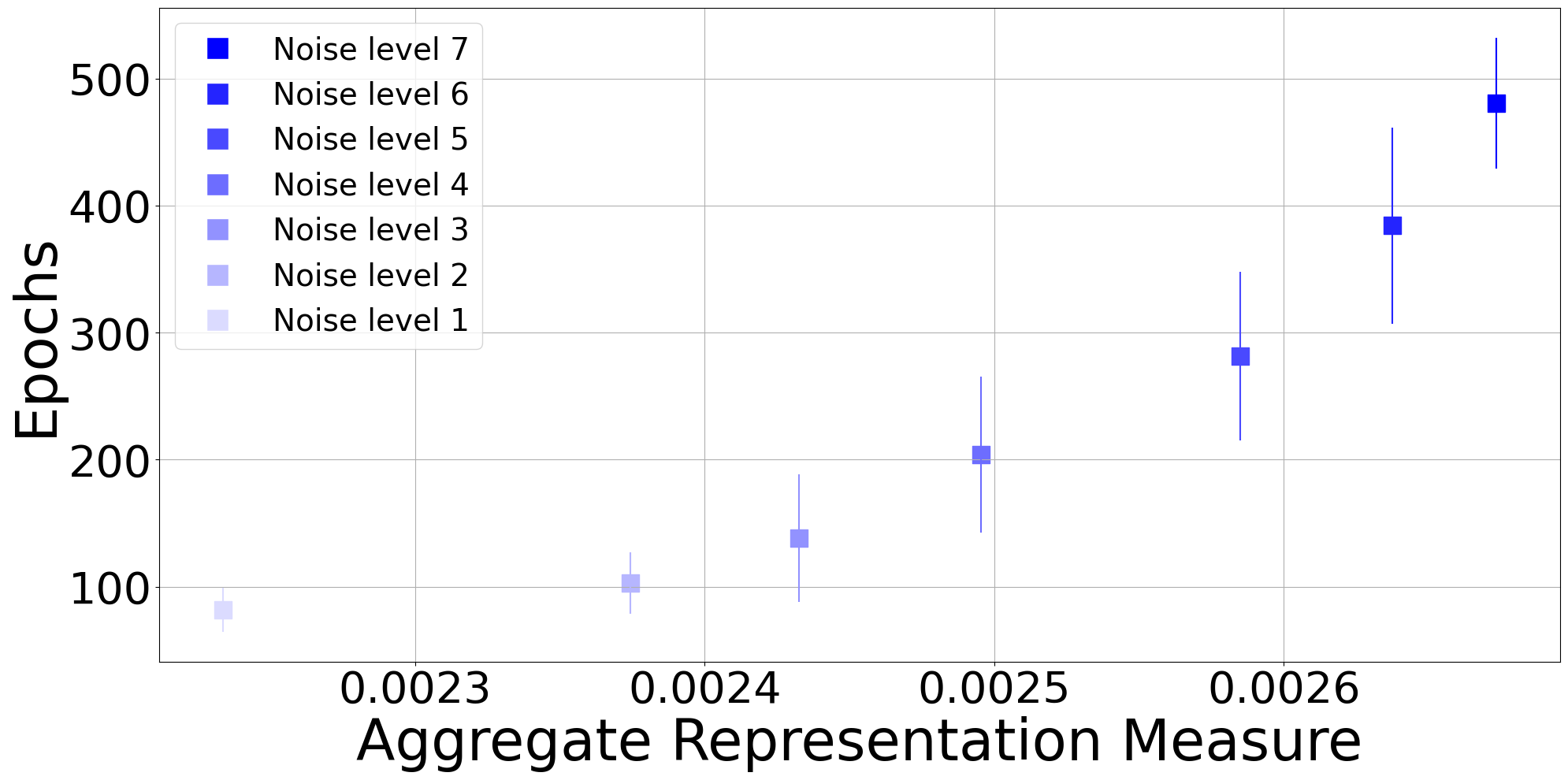}
        \caption{GoogLeNet - Gaussian noise}
    \end{subfigure}
    \hfill
    \begin{subfigure}{0.45\textwidth}
        \includegraphics[width=\linewidth]{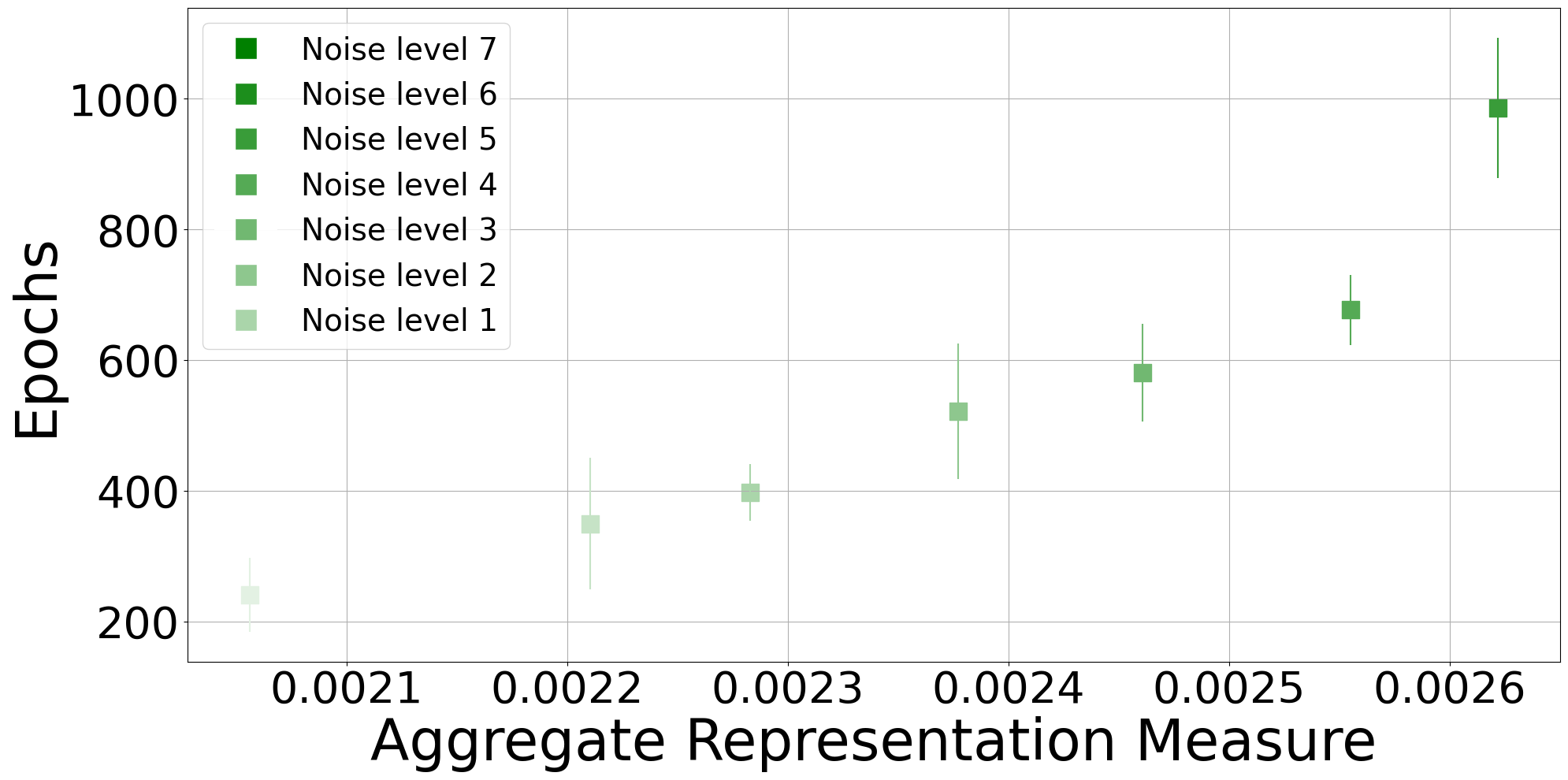}
        \caption{ResNet18 - Gaussian noise}
    \end{subfigure}
    \begin{subfigure}{0.45\textwidth}
        \includegraphics[width=\linewidth]{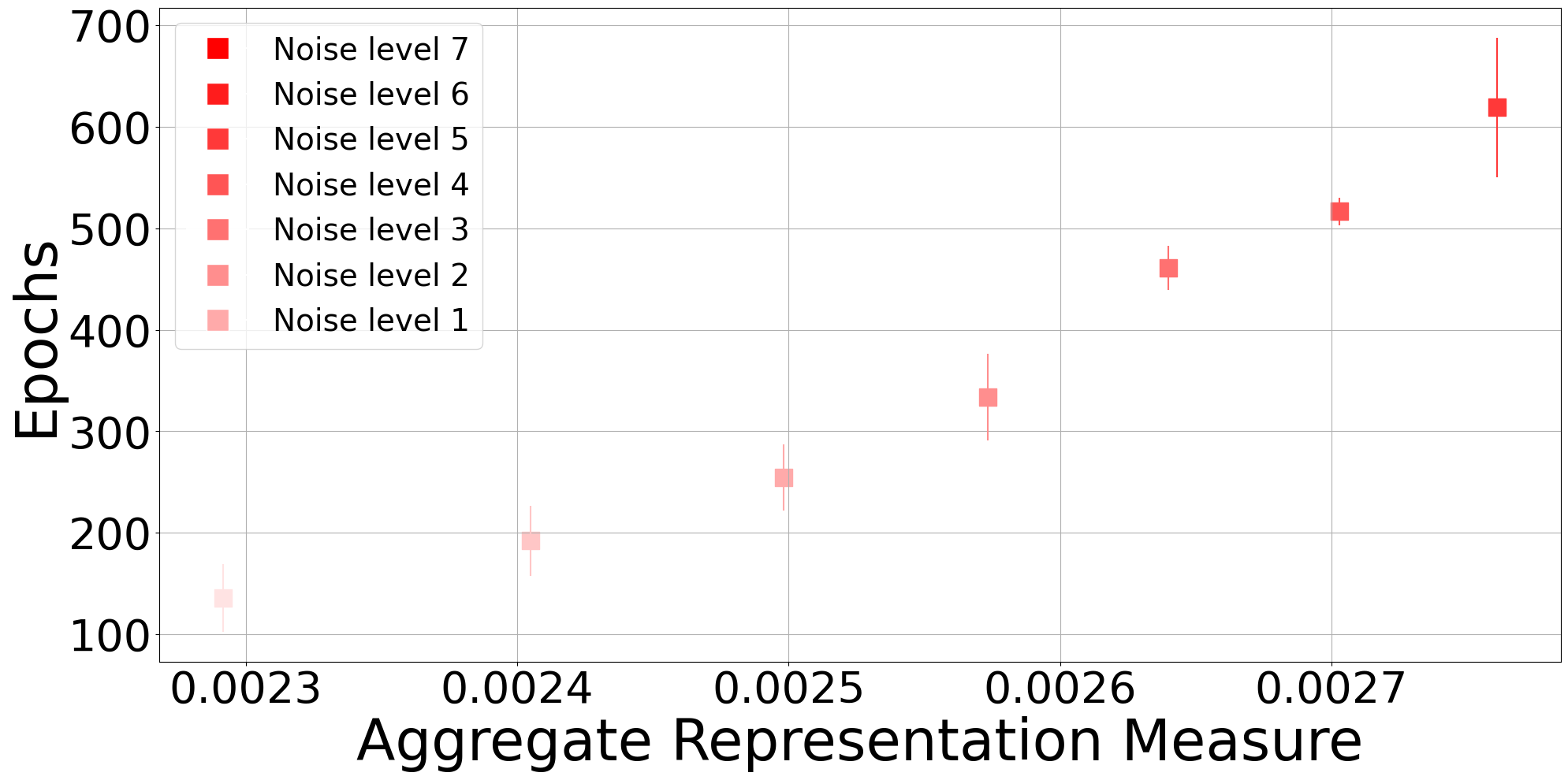}
        \caption{MobileNetv2 - Gaussian noise}
    \end{subfigure}
    \hfill
    \begin{subfigure}{0.45\textwidth}
        \includegraphics[width=\linewidth]{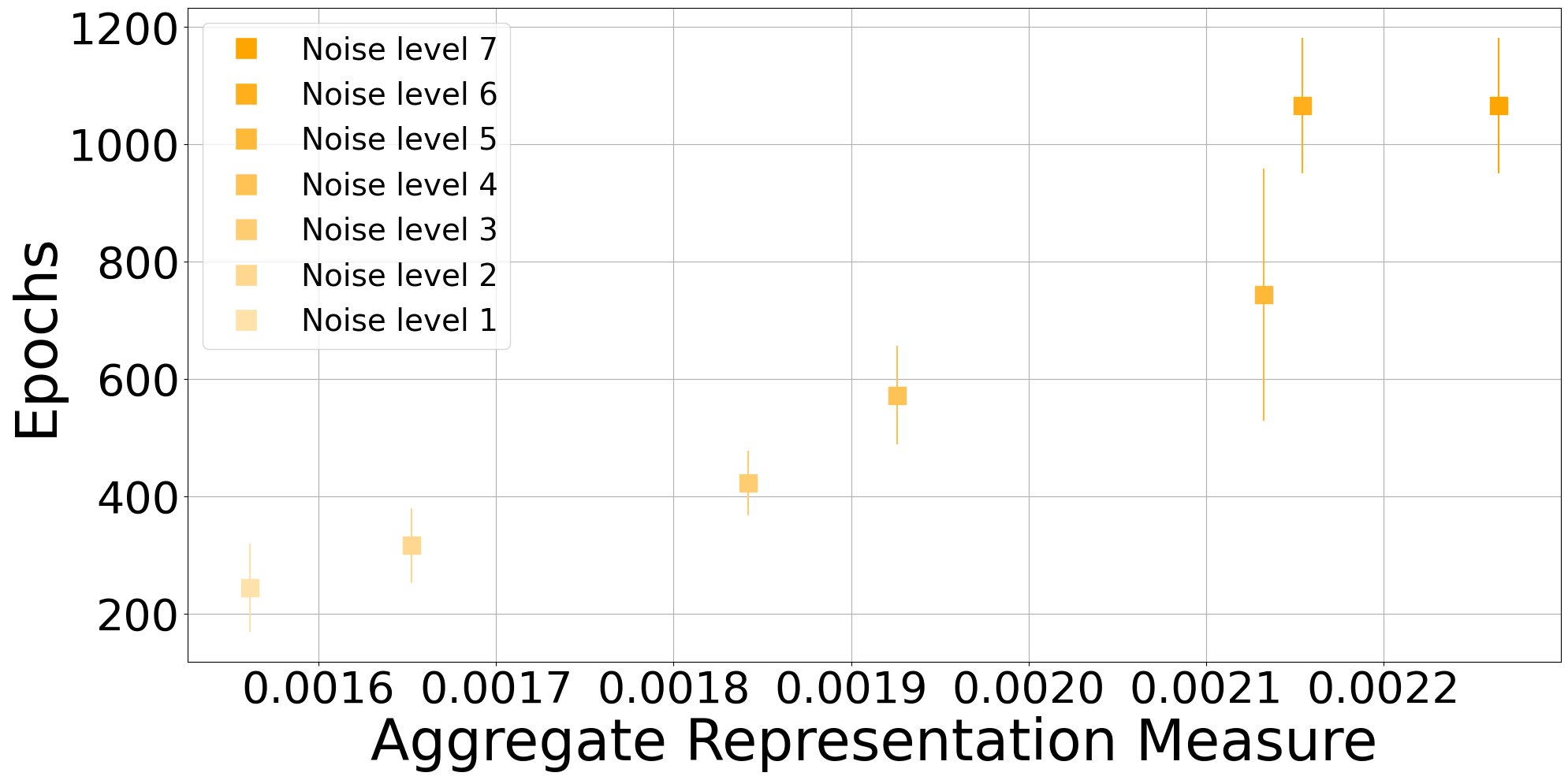}
        \caption{VGG16 - Gaussian noise}
    \end{subfigure}
    \caption{ARM vs Retraining Epochs on CIFAR10 dataset with Gaussian noise}
    \label{fig:CIFAR10_Gauss}
\end{figure}

\begin{figure}[ht]
    \centering
    \begin{subfigure}{0.45\textwidth}
        \includegraphics[width=\linewidth]{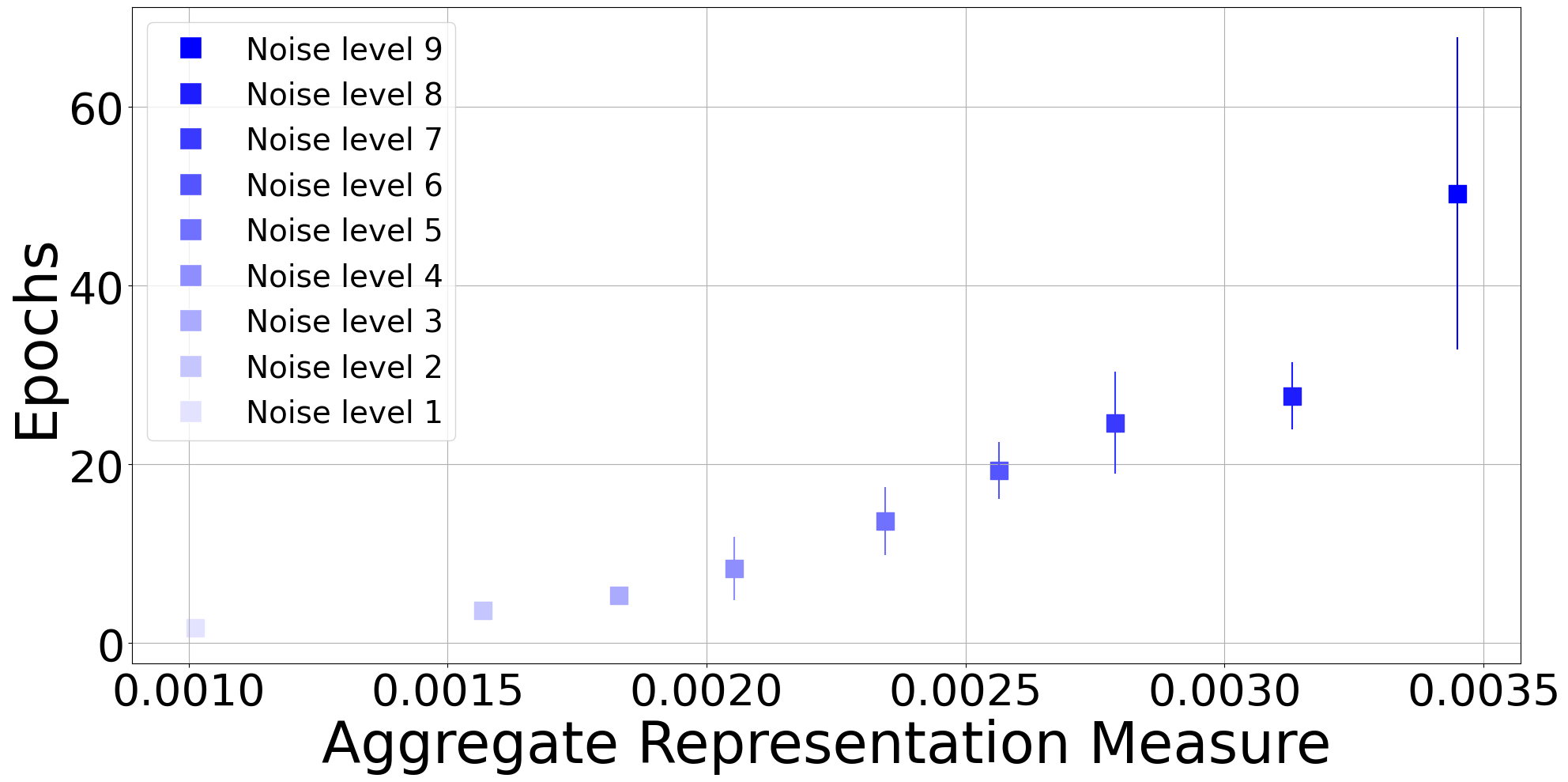}
        \caption{GoogLeNet - Salt-and-Pepper noise}
    \end{subfigure}
    \hfill
    \begin{subfigure}{0.45\textwidth}
        \includegraphics[width=\linewidth]{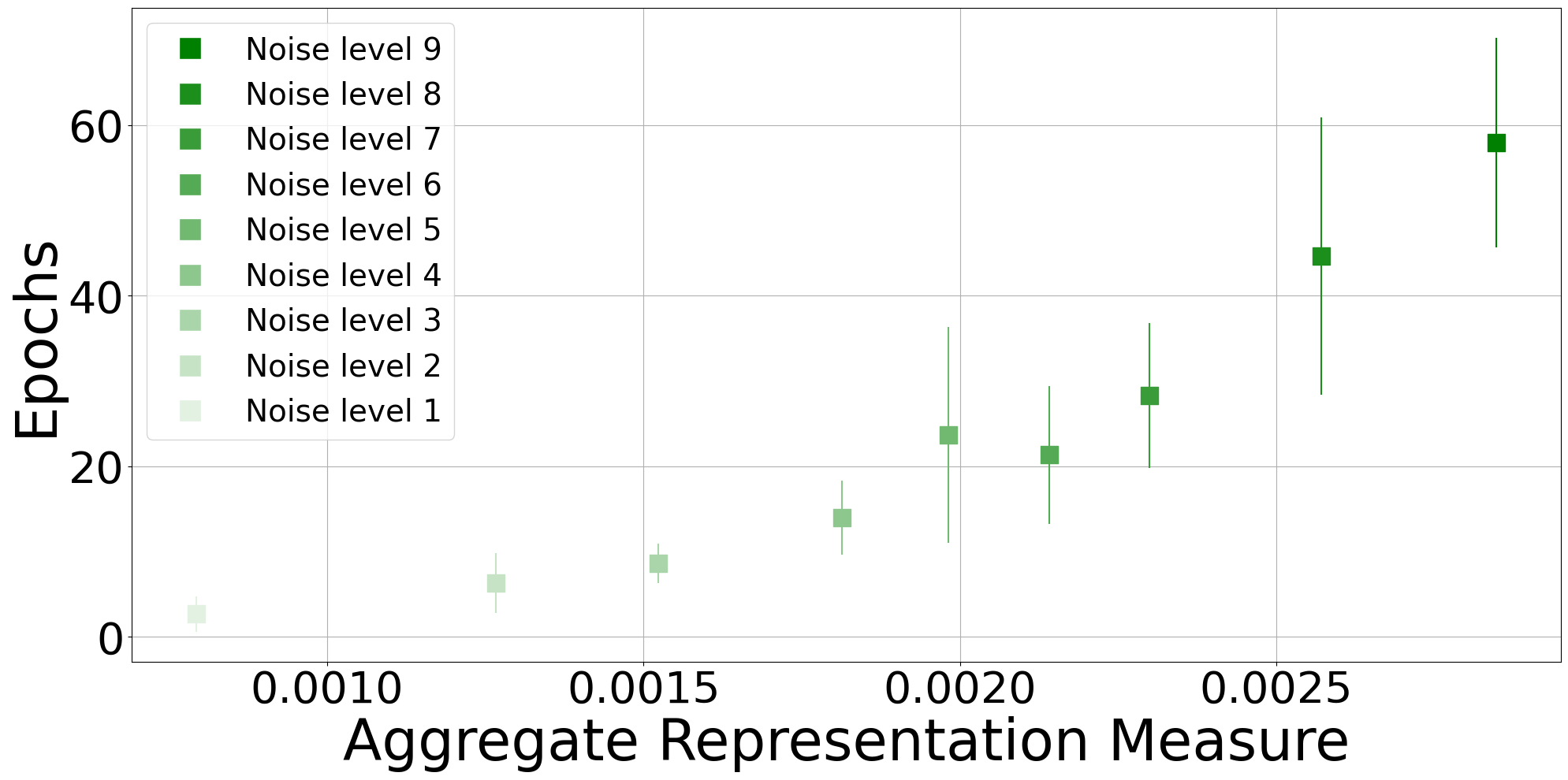}
        \caption{ResNet18 - Salt-and-Pepper noise}
    \end{subfigure}
    \begin{subfigure}{0.45\textwidth}
        \includegraphics[width=\linewidth]{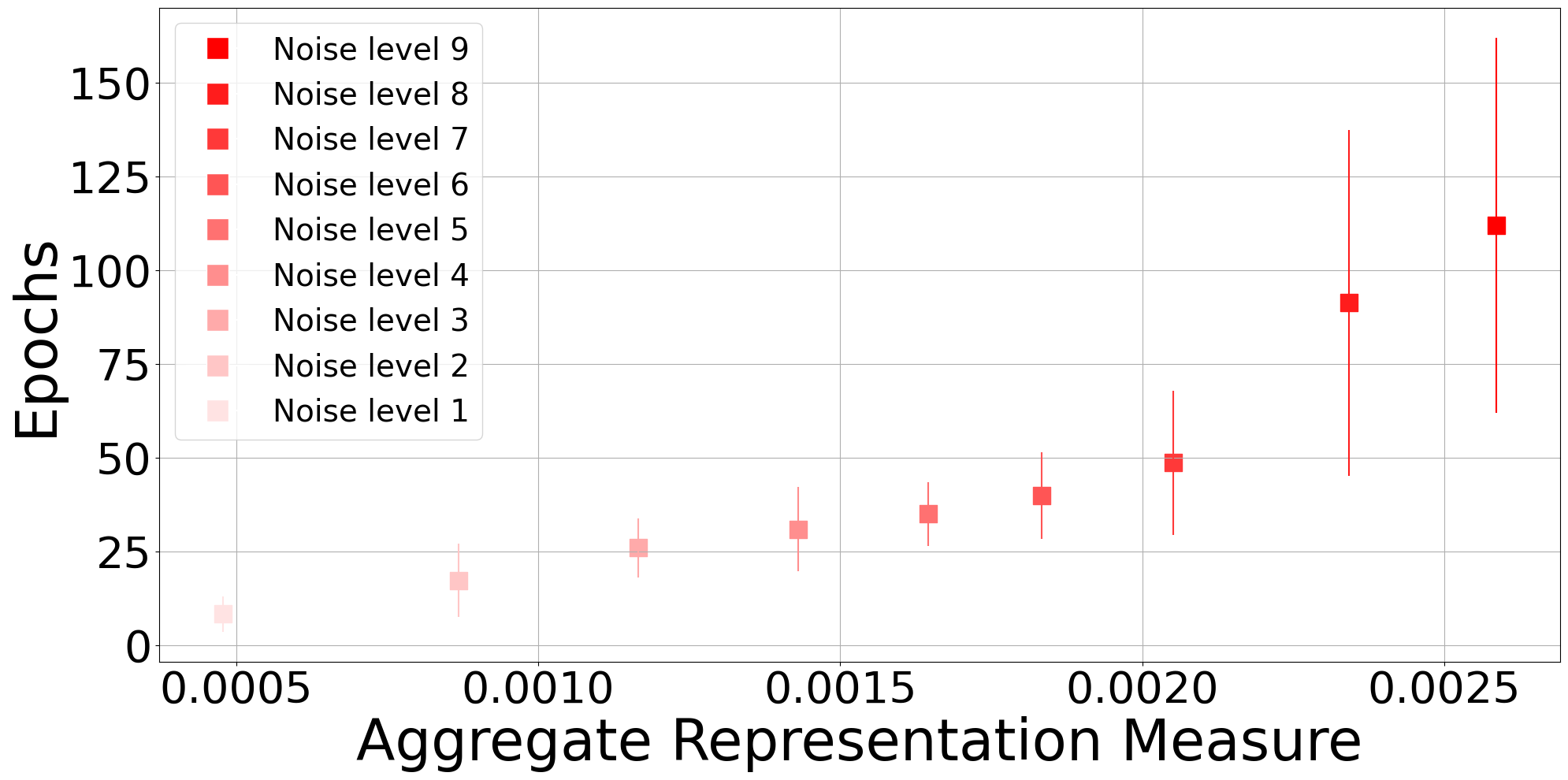}
        \caption{MobileNetv2 - Salt-and-Pepper noise}
    \end{subfigure}
    \hfill
    \begin{subfigure}{0.45\textwidth}
        \includegraphics[width=\linewidth]{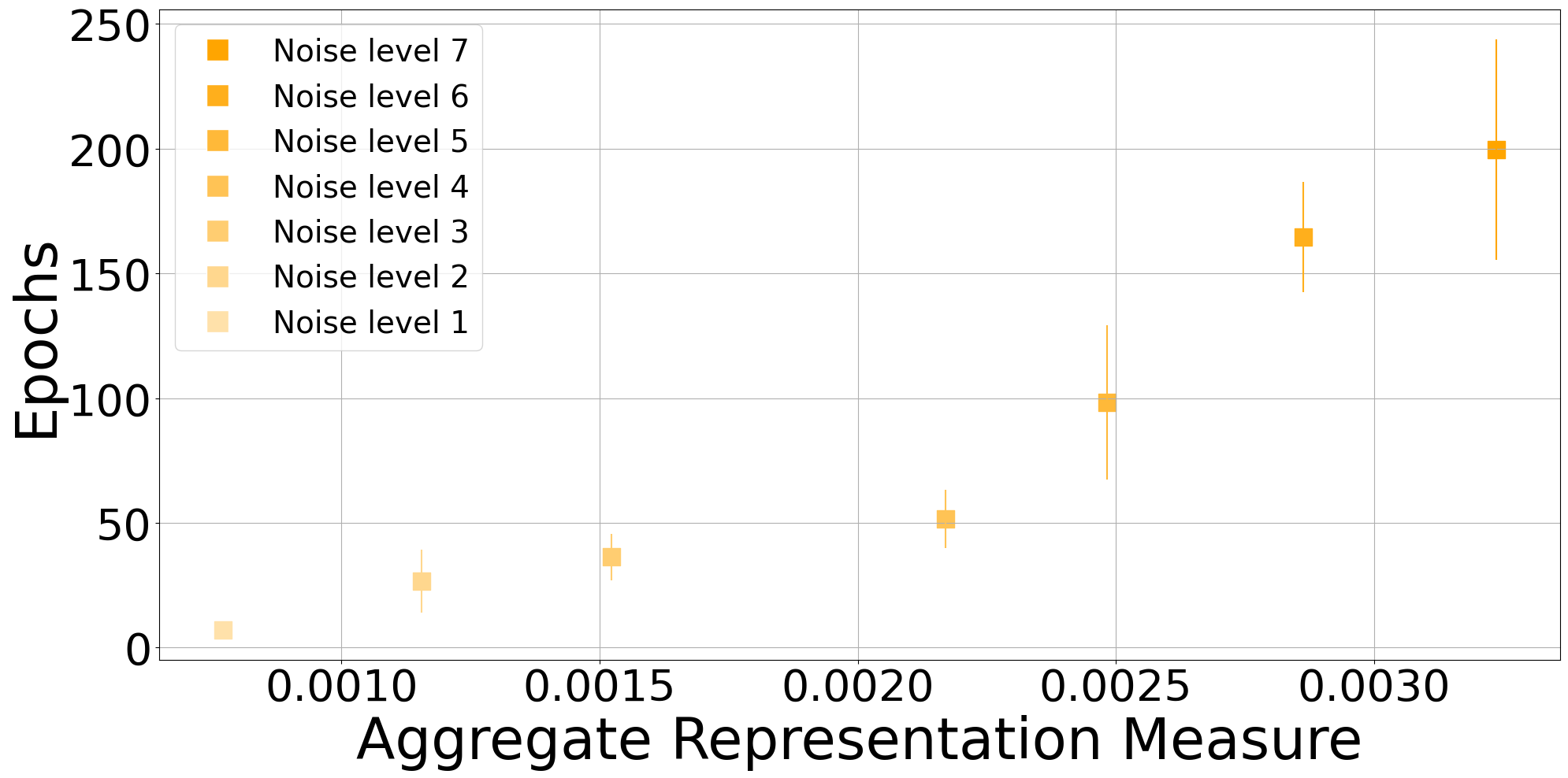}
        \caption{VGG16 - Salt-and-Pepper noise}
    \end{subfigure}
    \caption{ARM vs Retraining Epochs on CIFAR10 dataset with Salt and Pepper noise}
    \label{fig:CIFAR10_SaltPepper}
\end{figure}

\begin{figure}[ht]
    \centering
    \begin{subfigure}{0.45\textwidth}
        \includegraphics[width=\linewidth]{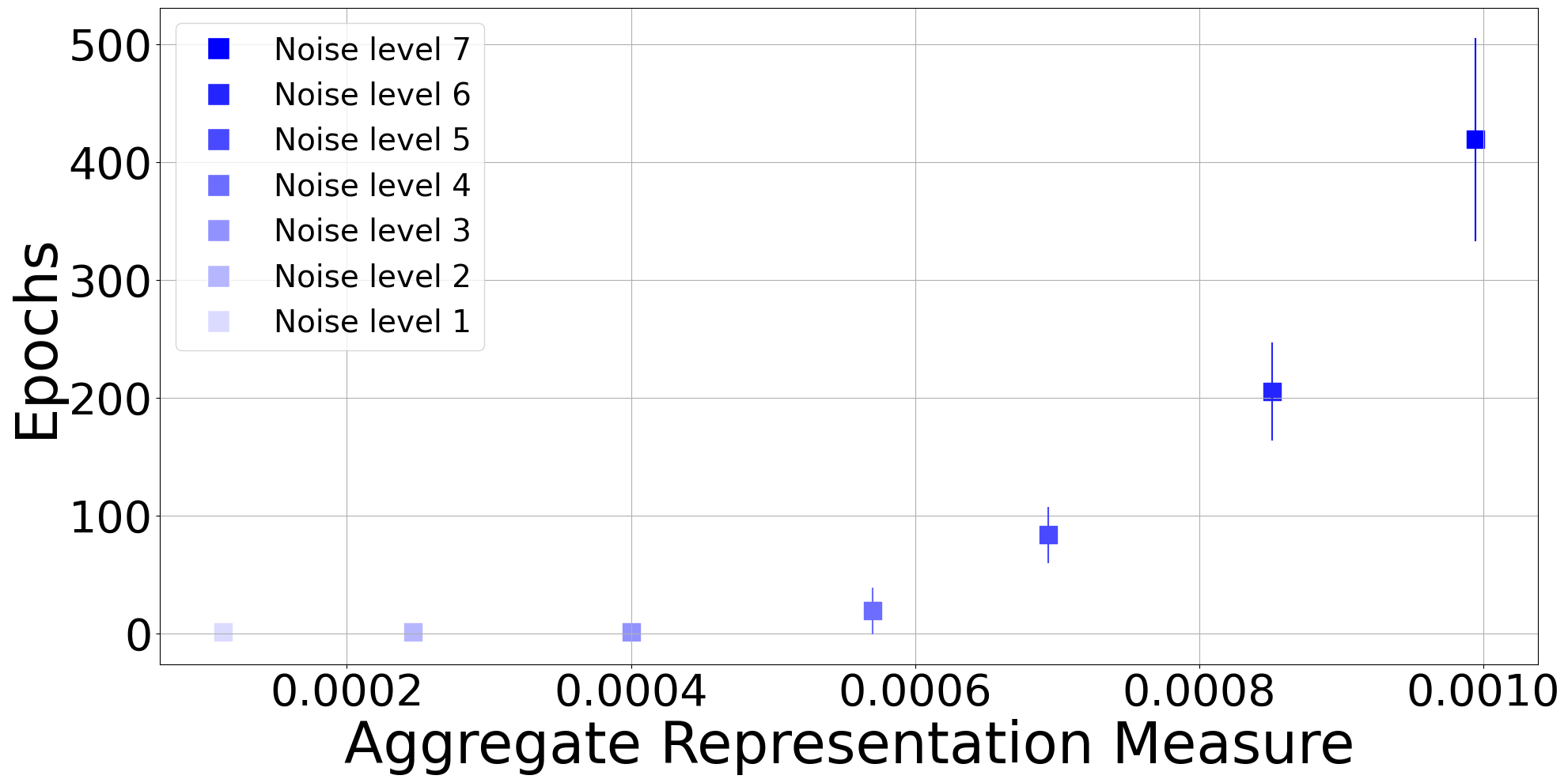}
        \caption{GoogLeNet - Image Blur}
    \end{subfigure}
    \hfill
    \begin{subfigure}{0.45\textwidth}
        \includegraphics[width=\linewidth]{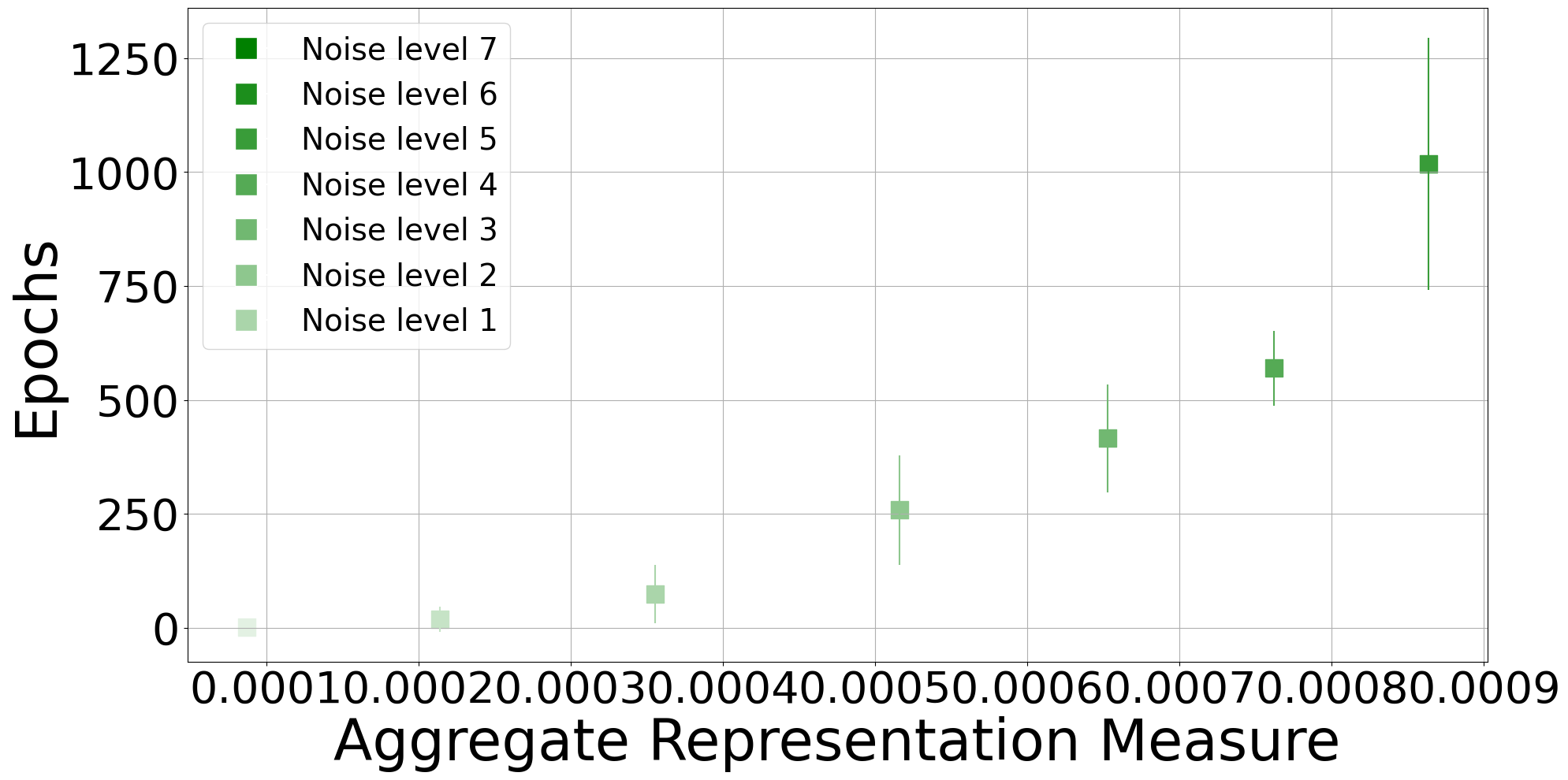}
        \caption{ResNet18 - Image Blur}
    \end{subfigure}
    \begin{subfigure}{0.45\textwidth}
        \includegraphics[width=\linewidth]{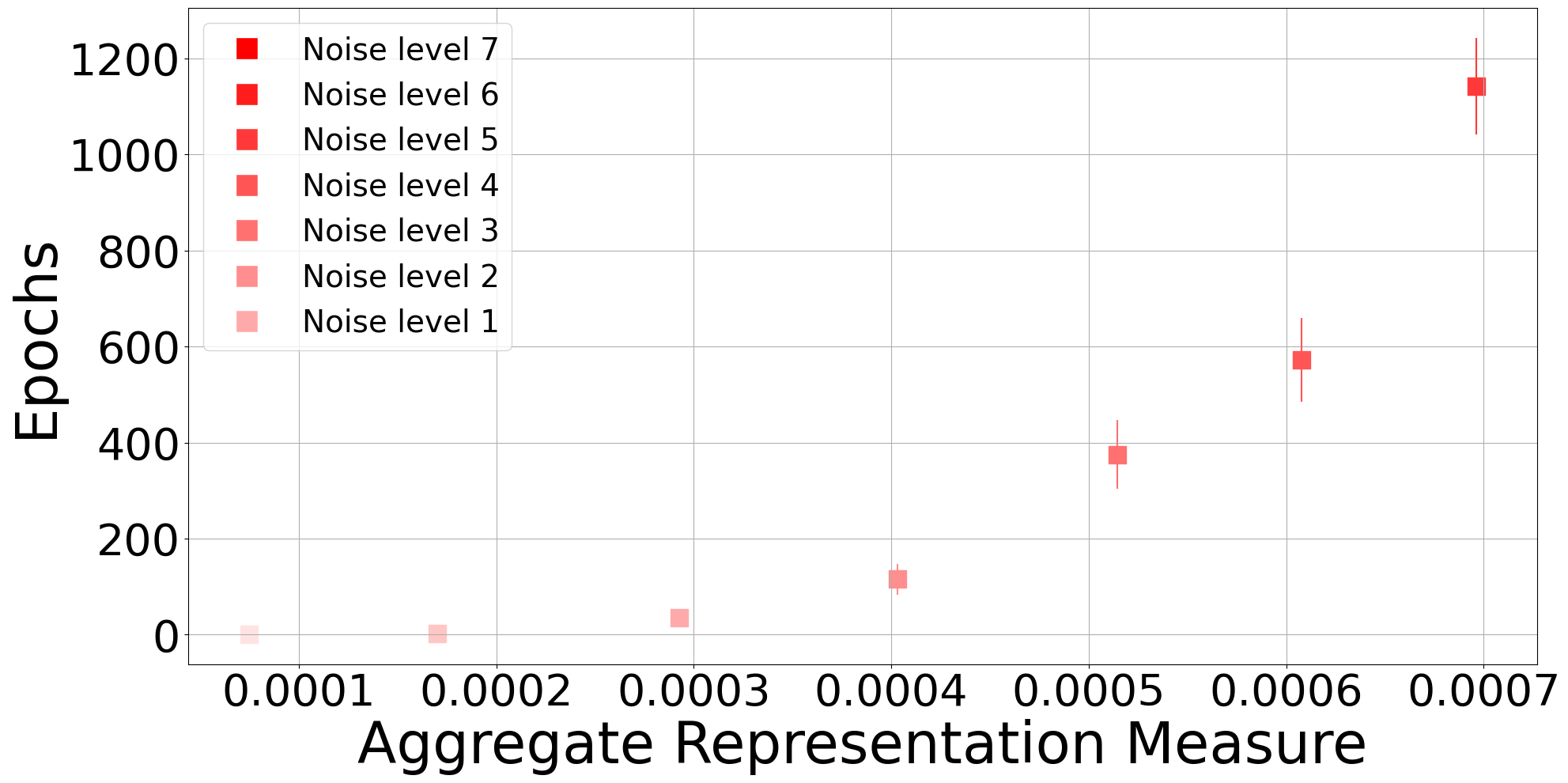}
        \caption{MobileNetv2 - Image Blur}
    \end{subfigure}
    \hfill
    \begin{subfigure}{0.45\textwidth}
        \includegraphics[width=\linewidth]{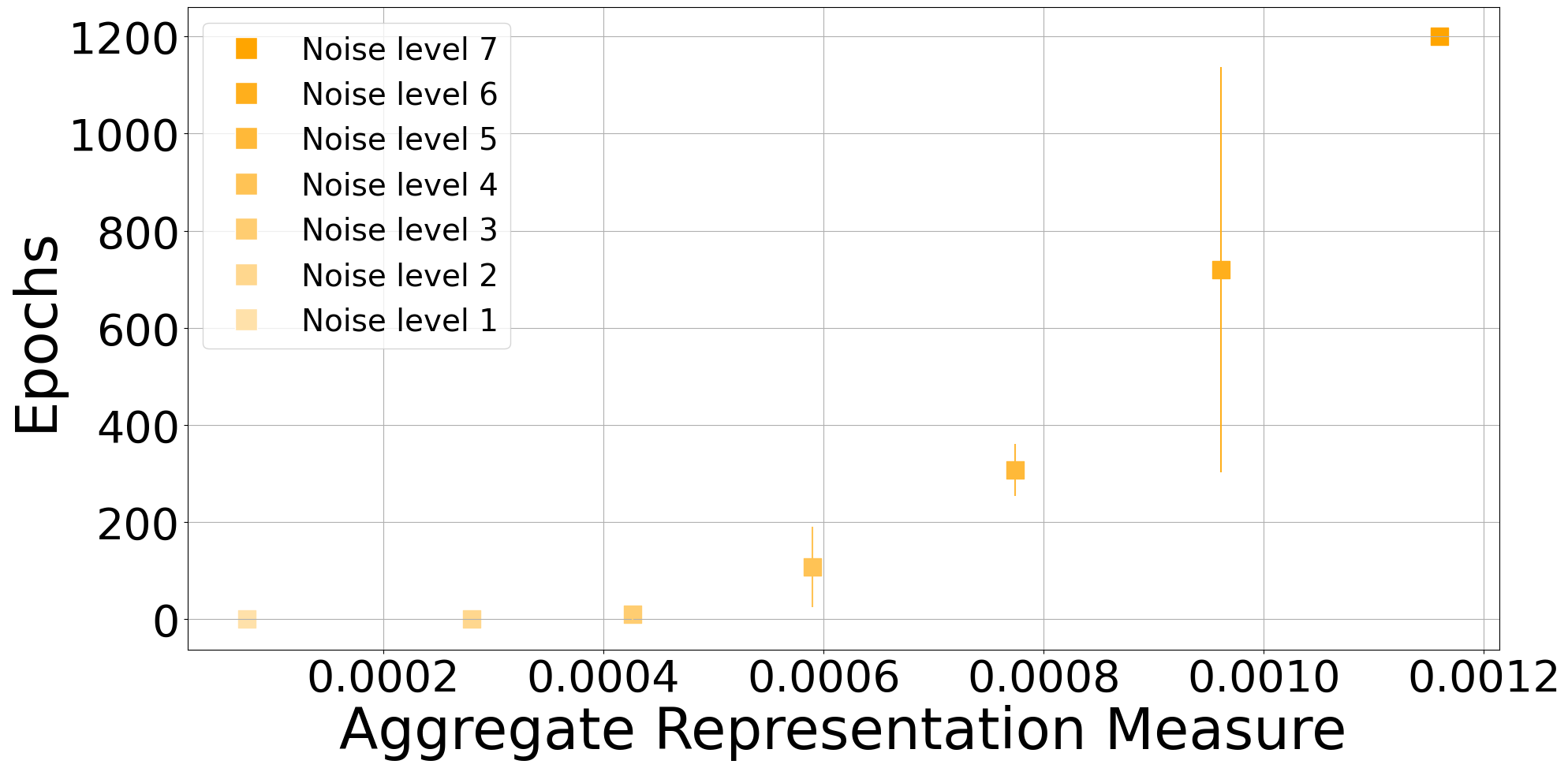}
        \caption{VGG16 - Image Blur}
    \end{subfigure}
    \caption{ARM vs Retraining Epochs on CIFAR10 dataset with Image Blur}
    \label{fig:CIFAR10_ImageBlur}
\end{figure}

This section provides all results for GoogleNet, ResNet18, MobileNetV2 and VGG16 on the CIFAR10 dataset. Fig.~\ref{fig:CIFAR10_Gauss}, Fig.~\ref{fig:CIFAR10_SaltPepper}, and Fig.~\ref{fig:CIFAR10_ImageBlur} illustrates the ARM and retraining epochs for different levels of Gaussian noise, Salt-and-Pepper noise and Image Blur, respectively. Table.~\ref{tab:correlation_table_C10} provides the Pearson correlation coefficient and p-values for the 4 models retrained on CIFAR10 for the 3 noise types.

\begin{table}[htbp]
\small
        \centering
        \begin{tabular}{ccccccc}
             \toprule
             Model & Coefficient & p-value & Coefficient & p-value & Coefficient & p-value \\
             \midrule
             \multicolumn{1}{c}{ } &\multicolumn{2}{c}{Salt-Pepper} & \multicolumn{2}{c}{Gaussian} & \multicolumn{2}{c}{Blur} \\
             \midrule
             GoogLeNet & 0.92 & 0.00041 & 0.93 & 0.002 & 0.85 & 0.015 \\
             \hline
             ResNet18 & 0.93 & 0.00024 & 0.94 & 0.0014 & 0.92 & 0.0030 \\
             \hline
             MobileNetV2 & 0.91 & 0.00061 & 0.97 & 0.00016 & 0.87 & 0.0091 \\
             \hline
             VGG16 & 0.94 & 0.0014 &  0.95 & 0.0008 & 0.90 & 0.0051 \\
             \bottomrule
        \end{tabular}
        \vspace{0.1cm}
        \caption{Pearson correlation between epochs and ARM - CIFAR10 Dataset}
        \label{tab:correlation_table_C10}
\end{table}

\subsection{SVHN Dataset}
\begin{figure}[ht]
    \centering
    \begin{subfigure}{0.45\textwidth}
        \includegraphics[width=\linewidth]{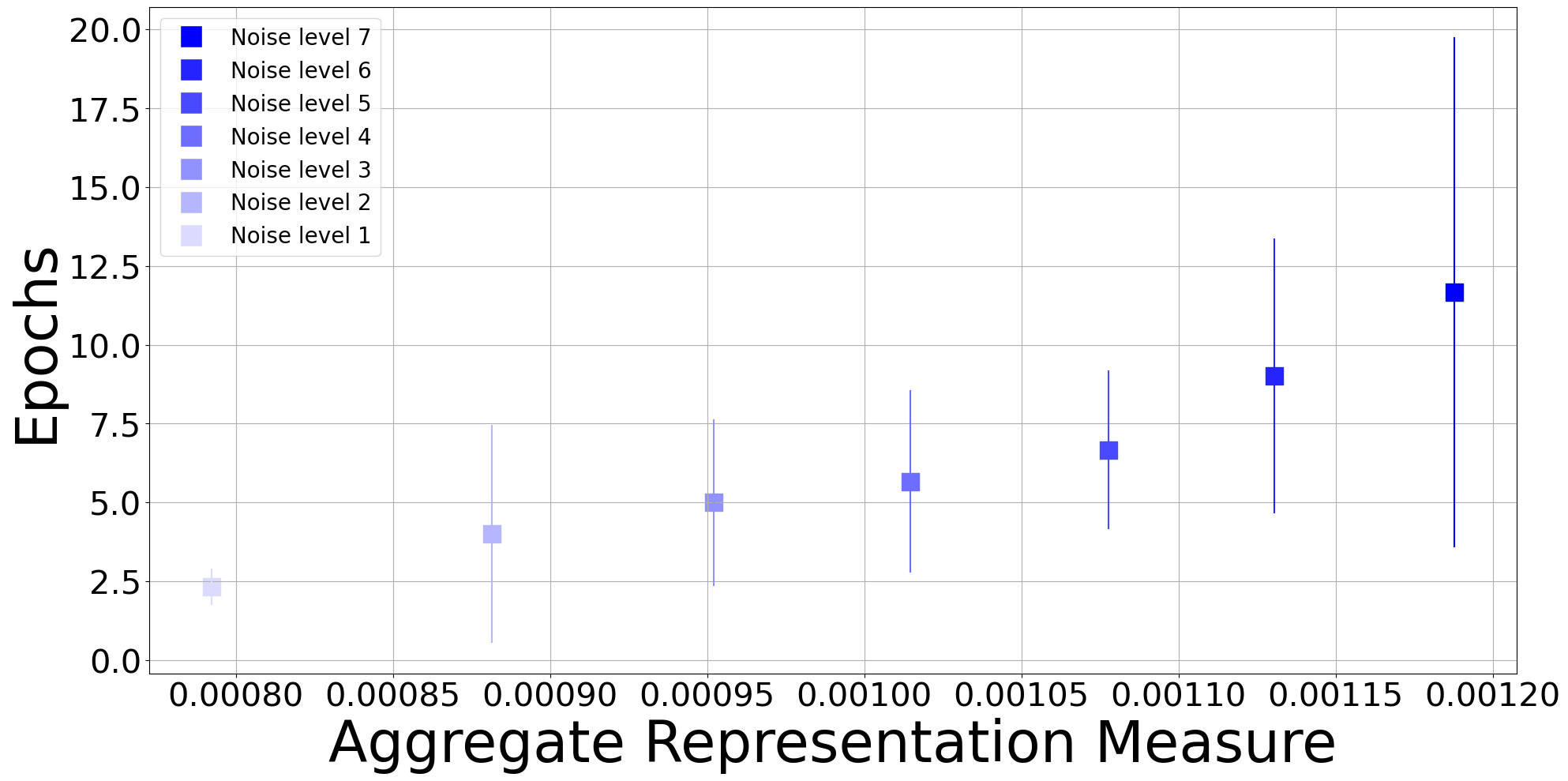}
        \caption{GoogLeNet - Gaussian noise}
    \end{subfigure}
    \hfill
    \begin{subfigure}{0.45\textwidth}
        \includegraphics[width=\linewidth]{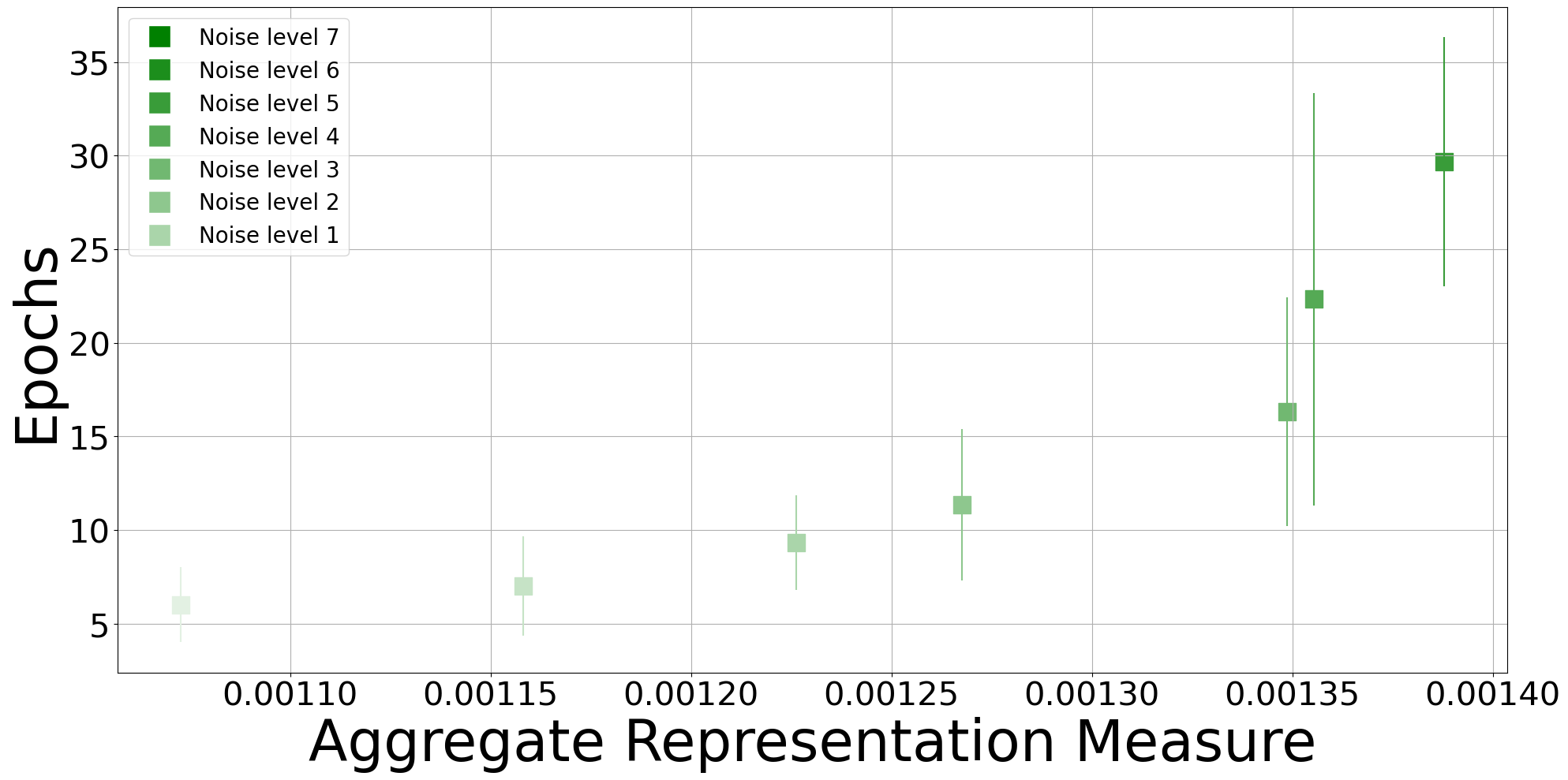}
        \caption{ResNet18 - Gaussian noise}
    \end{subfigure}
    \begin{subfigure}{0.45\textwidth}
        \includegraphics[width=\linewidth]{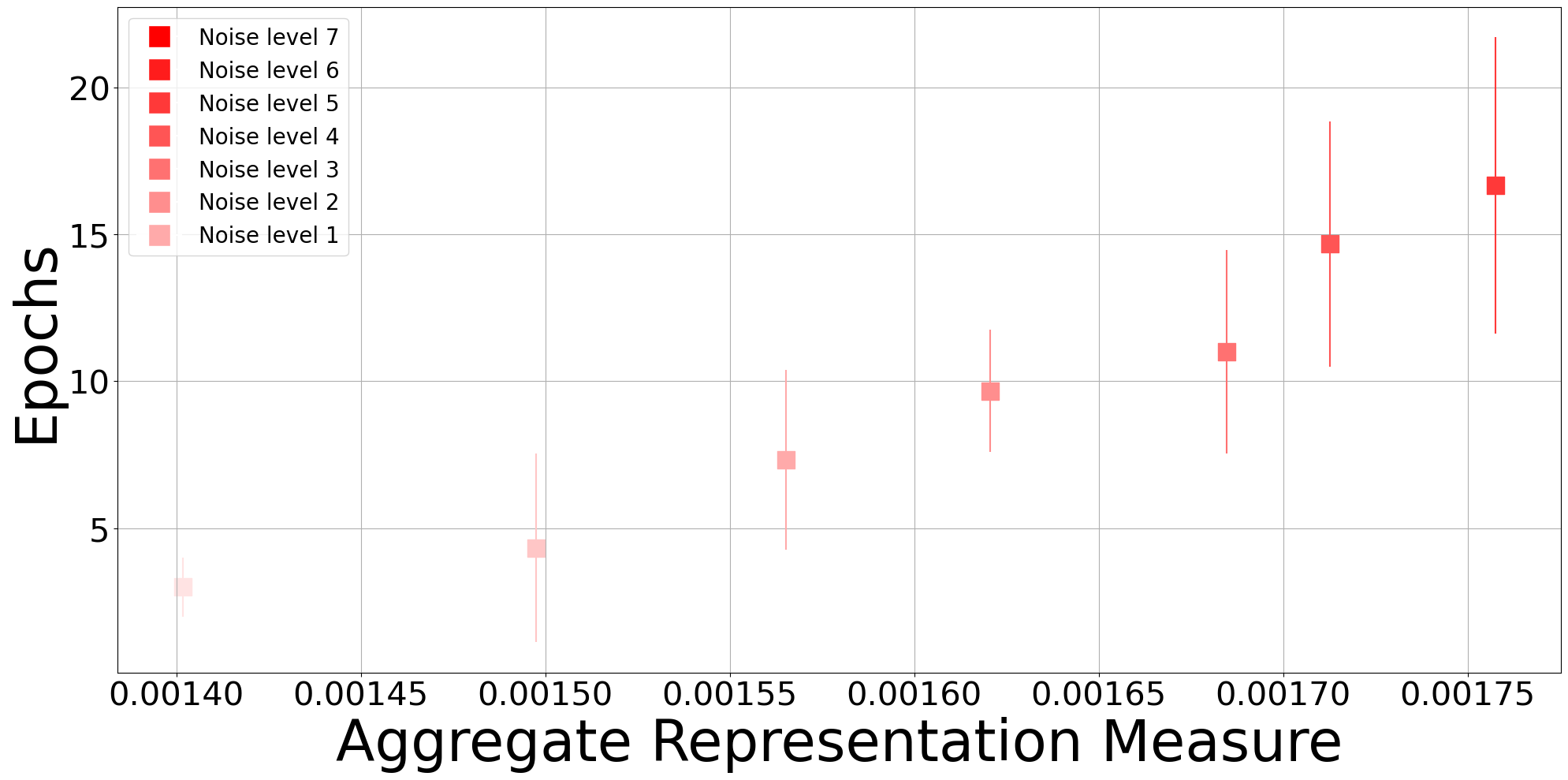}
        \caption{MobileNetv2 - Gaussian noise}
    \end{subfigure}
    \hfill
    \begin{subfigure}{0.45\textwidth}
        \includegraphics[width=\linewidth]{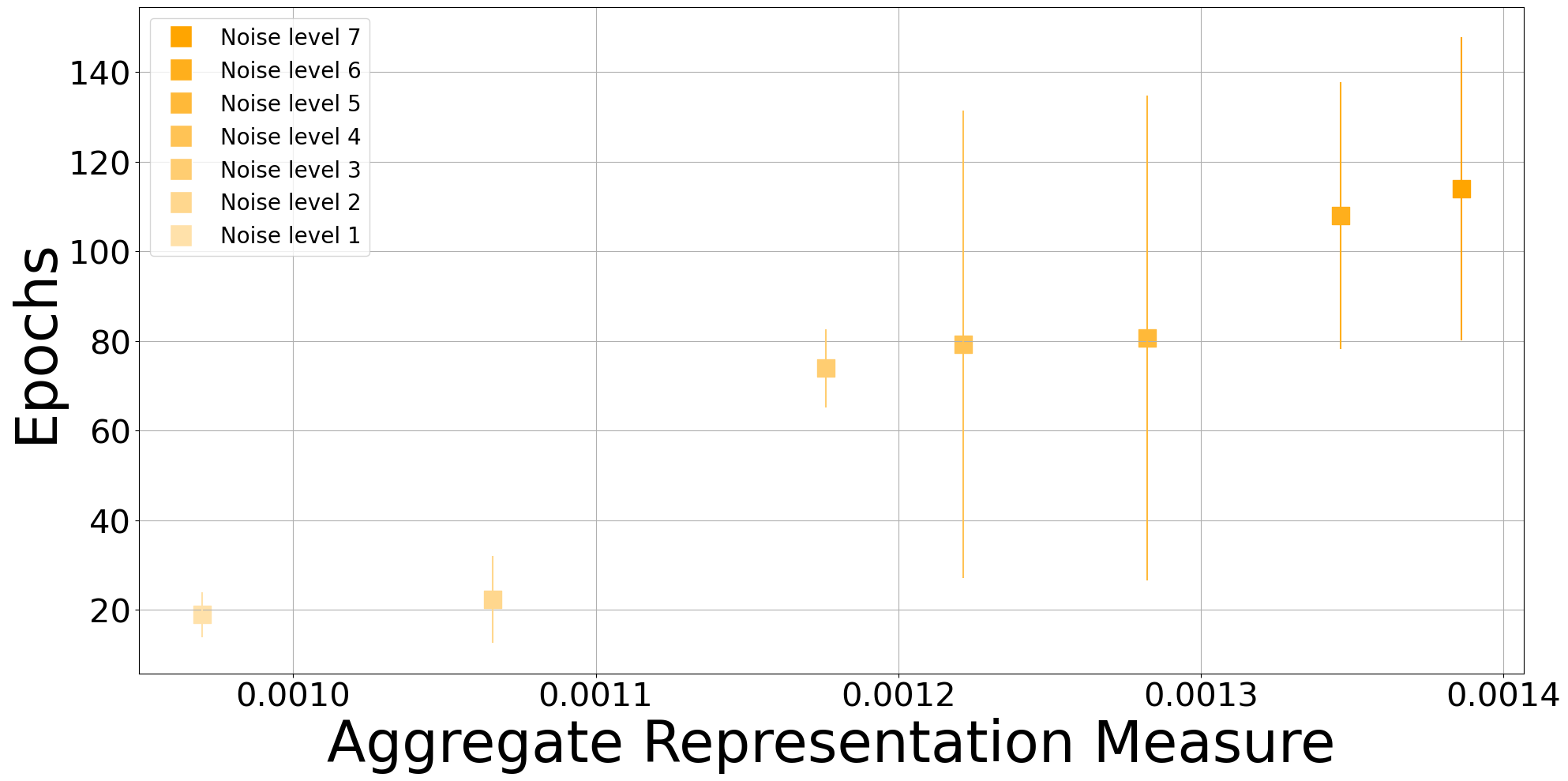}
        \caption{VGG16 - Gaussian noise}
    \end{subfigure}
    \caption{ARM vs Retraining Epochs on SVHN dataset with Gaussian noise}
    \label{fig:SVHN_Gauss}
\end{figure}

\begin{figure}[ht]
    \centering
    \begin{subfigure}{0.45\textwidth}
        \includegraphics[width=\linewidth]{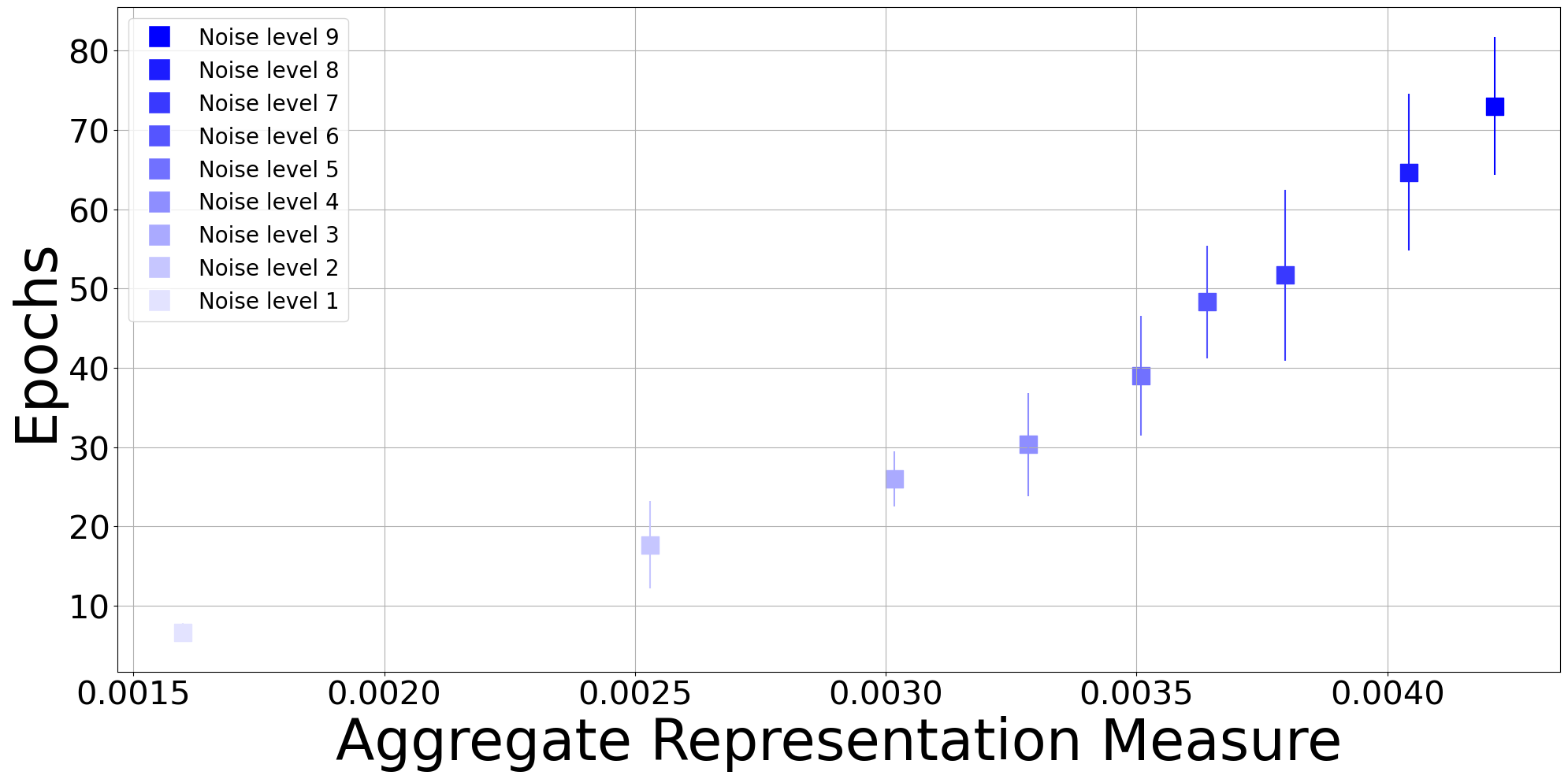}
        \caption{GoogLeNet - Salt-and-Pepper noise}
    \end{subfigure}
    \hfill
    \begin{subfigure}{0.45\textwidth}
        \includegraphics[width=\linewidth]{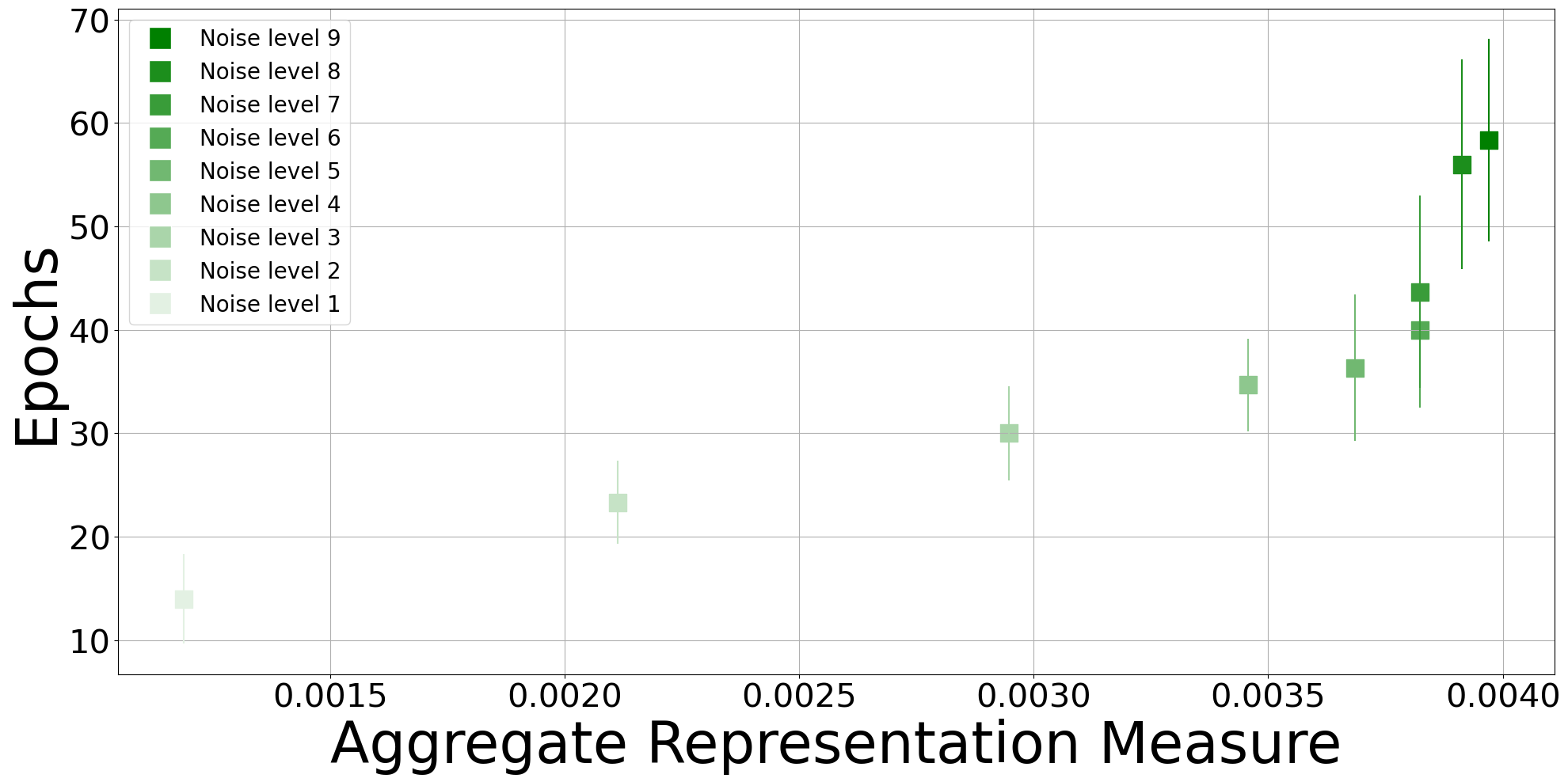}
        \caption{ResNet18 - Salt-and-Pepper noise}
    \end{subfigure}
    \begin{subfigure}{0.45\textwidth}
        \includegraphics[width=\linewidth]{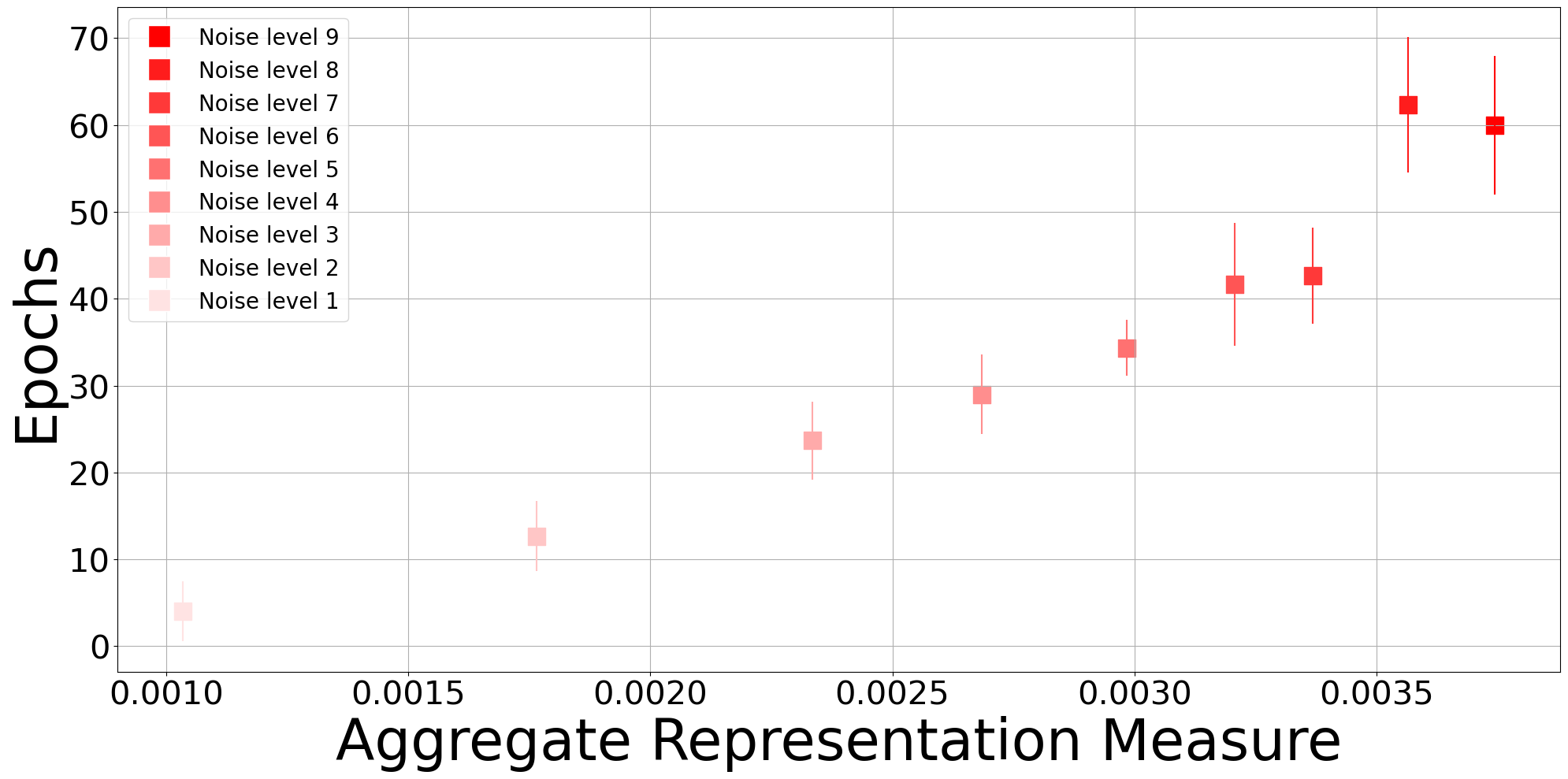}
        \caption{MobileNetv2 - Salt-and-Pepper noise}
    \end{subfigure}
    \hfill
    \begin{subfigure}{0.45\textwidth}
        \includegraphics[width=\linewidth]{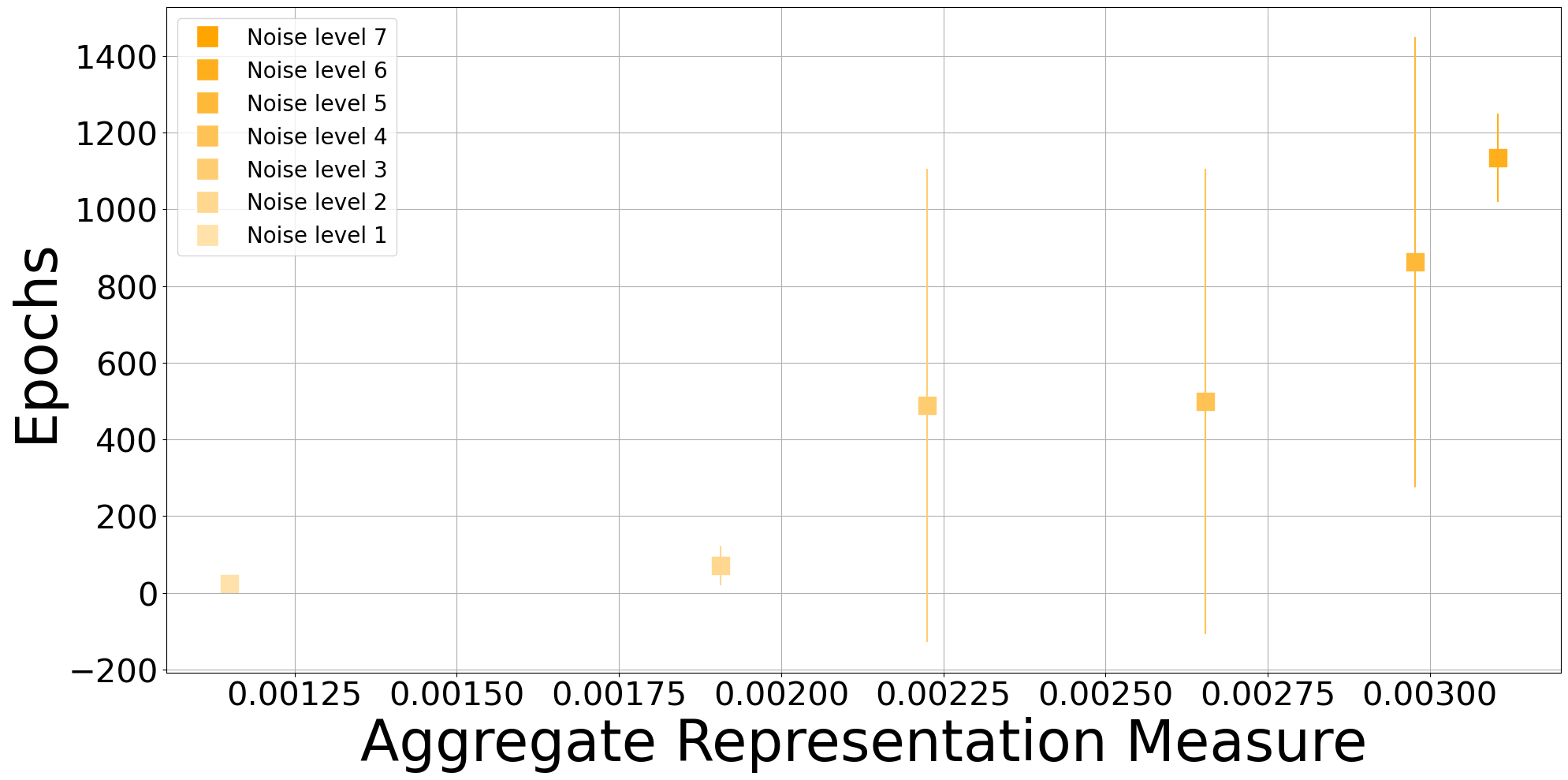}
        \caption{VGG16 - Salt-and-Pepper noise}
    \end{subfigure}
    \caption{ARM vs Retraining Epochs on SVHN dataset with Salt-and-Pepper noise}
    \label{fig:SVHN_SaltPepper}
\end{figure}

\begin{figure}[ht]
    \centering
    \begin{subfigure}{0.45\textwidth}
        \includegraphics[width=\linewidth]{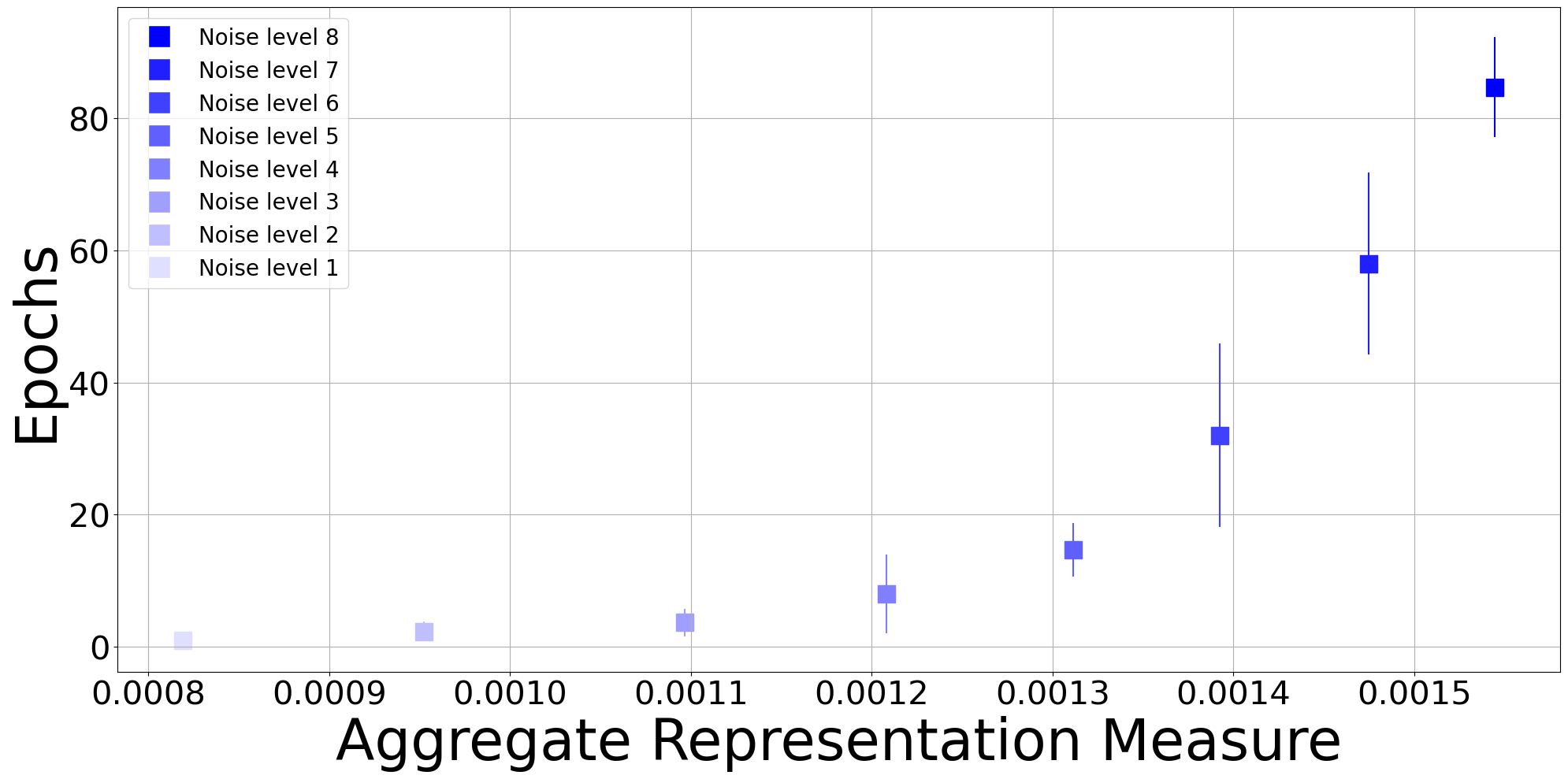}
        \caption{GoogLeNet - Image Blur}
    \end{subfigure}
    \hfill
    \begin{subfigure}{0.45\textwidth}
        \includegraphics[width=\linewidth]{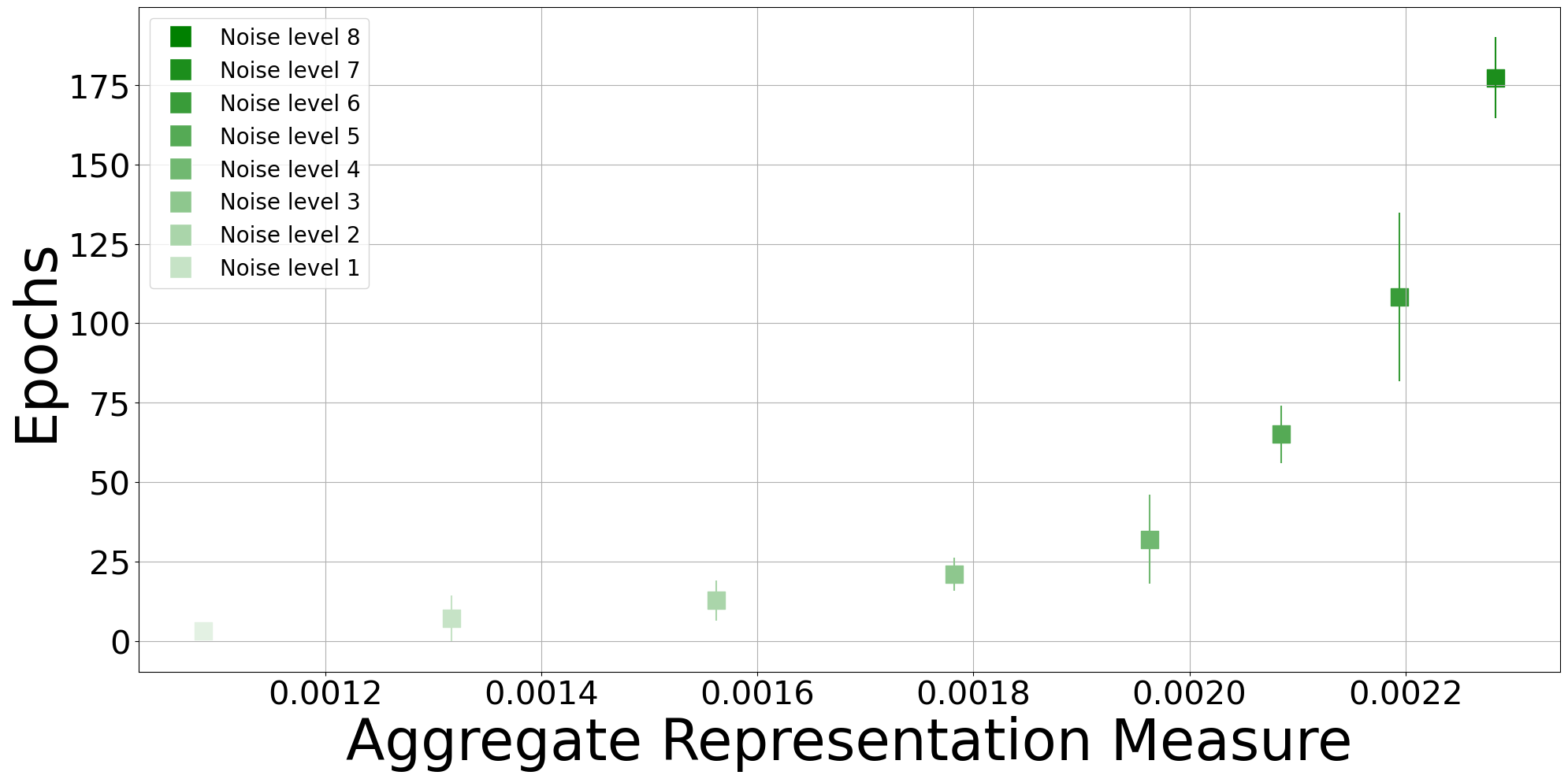}
        \caption{ResNet18 - Image Blur}
    \end{subfigure}
    \begin{subfigure}{0.45\textwidth}
        \includegraphics[width=\linewidth]{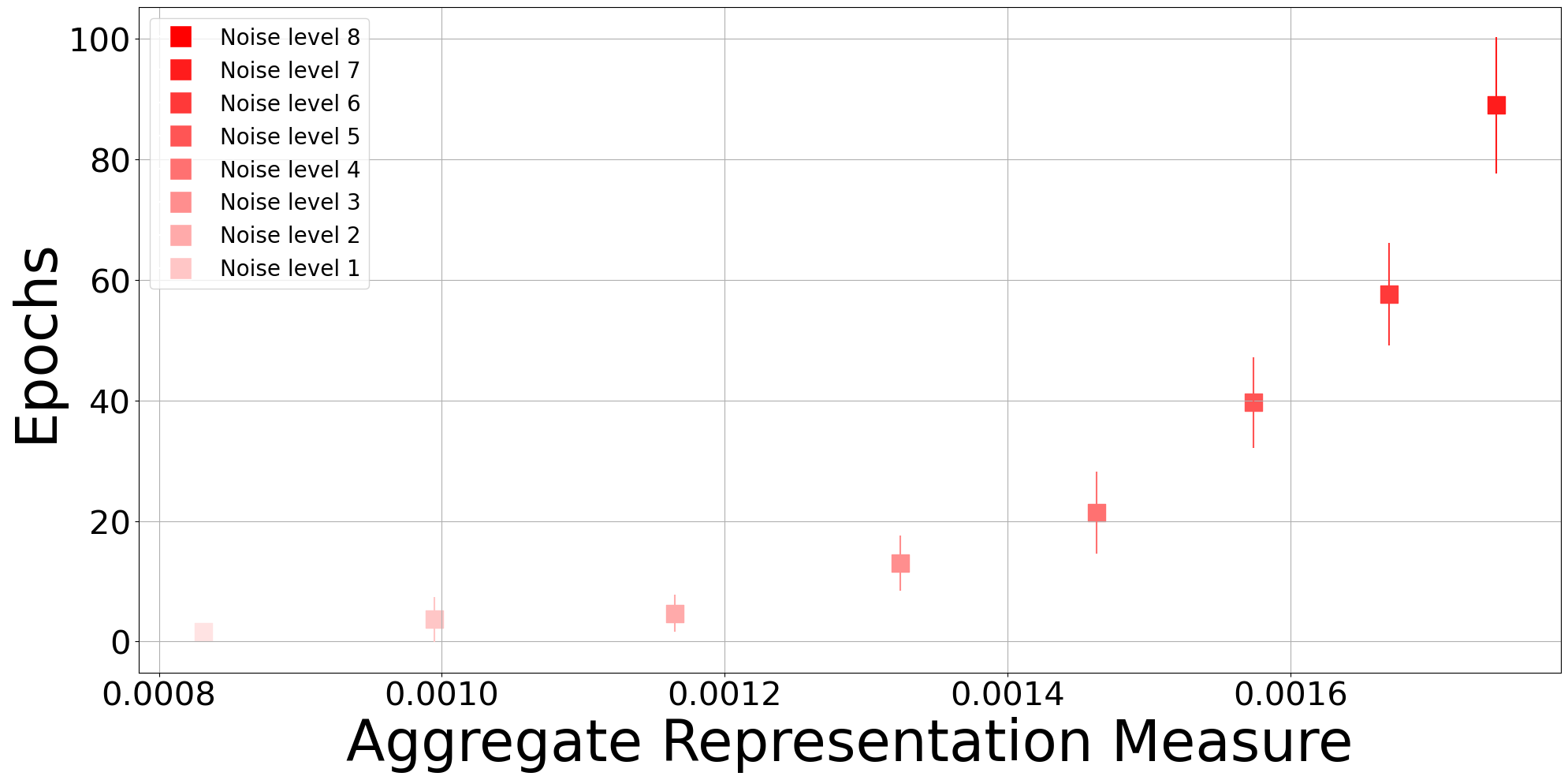}
        \caption{MobileNetv2 - Image Blur}
    \end{subfigure}
    \hfill
    \begin{subfigure}{0.45\textwidth}
        \includegraphics[width=\linewidth]{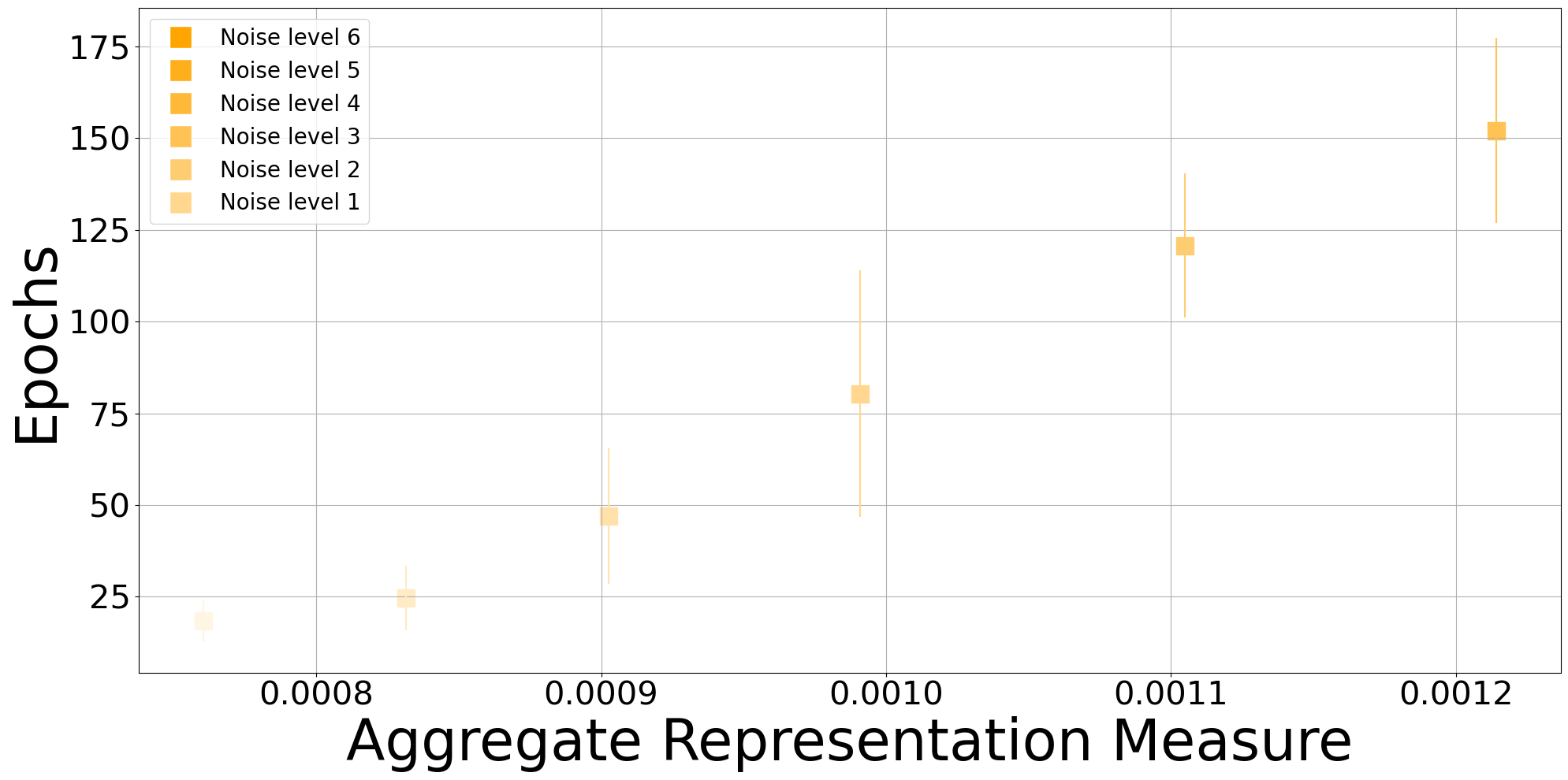}
        \caption{VGG16 - Image Blur}
    \end{subfigure}
    \caption{ARM vs Retraining Epochs on SVHN dataset with Image Blur}
    \label{fig:SVHN_Blur}
\end{figure}

This section provides all results for GoogleNet, ResNet18, MobileNetV2 and VGG16 on the SVHN dataset. Fig.~\ref{fig:SVHN_Gauss}, Fig.~\ref{fig:SVHN_SaltPepper}, and Fig.~\ref{fig:SVHN_Blur} illustrates the ARM and retraining epoch values for different levels of Gaussian noise, Salt-and-Pepper noise and Image Blur, respectively. Table.~\ref{tab:correlation_table_SVHN} provides the Pearson correlation coefficient and p-values for the 4 models retrained on SVHN for the 3 noise types.

\begin{table}[ht]
\small
        \centering
        \begin{tabular}{ccccccc}
             \toprule
             Model & Coefficient & p-value & Coefficient & p-value & Coefficient & p-value \\
             \midrule
             \multicolumn{1}{c}{ } &\multicolumn{2}{c}{Salt-Pepper} & \multicolumn{2}{c}{Gaussian} & \multicolumn{2}{c}{Blur} \\
             \midrule
             GoogLeNet & 0.94 & 0.00012 & 0.95 & 0.0006 & 0.84 & 0.0080 \\
             \hline
             ResNet18 & 0.88 & 0.0015 & 0.88 & 0.0084 & 0.811 & 0.014 \\
             \hline
             MobileNetV2 & 0.959 & 0.00041 & 0.97 & 0.00025 & 0.87 & 0.0047 \\
             \hline
             VGG16 & 0.91 & 0.0093 & 0.97 & 0.00023 & 0.99 & 0.000064 \\
             \bottomrule
        \end{tabular}
        \vspace{0.1cm}
        \caption{Pearson correlation between epochs and ARM - SVHN Dataset}
        \label{tab:correlation_table_SVHN}
\end{table}

\newpage

\subsection{CIFAR100 Dataset}
\begin{figure}[ht]
    \centering
    \begin{subfigure}{0.45\textwidth}
        \includegraphics[width=\linewidth]{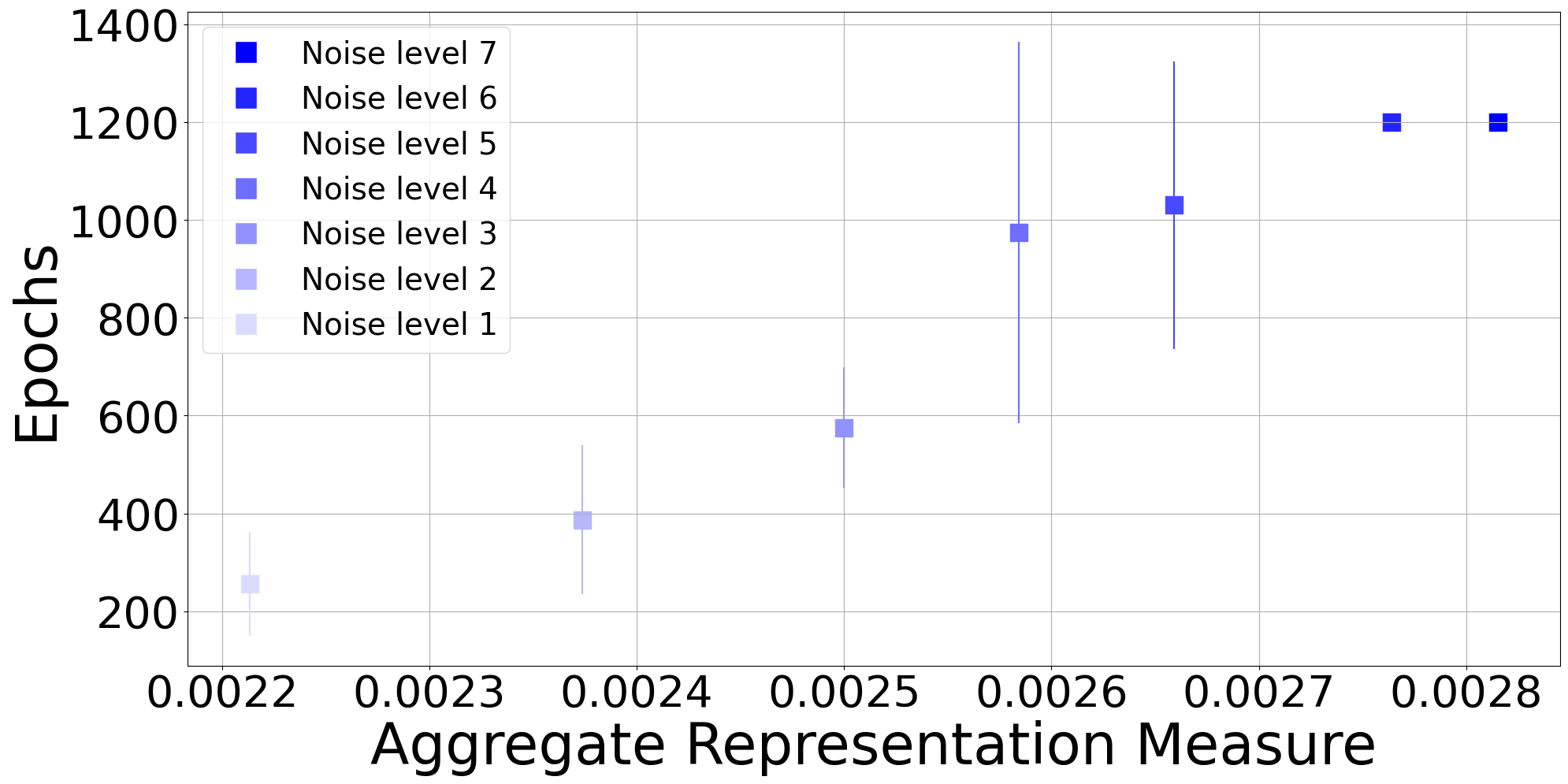}
        \caption{GoogLeNet - Gaussian noise}
    \end{subfigure}
    \hfill
    \begin{subfigure}{0.45\textwidth}
        \includegraphics[width=\linewidth]{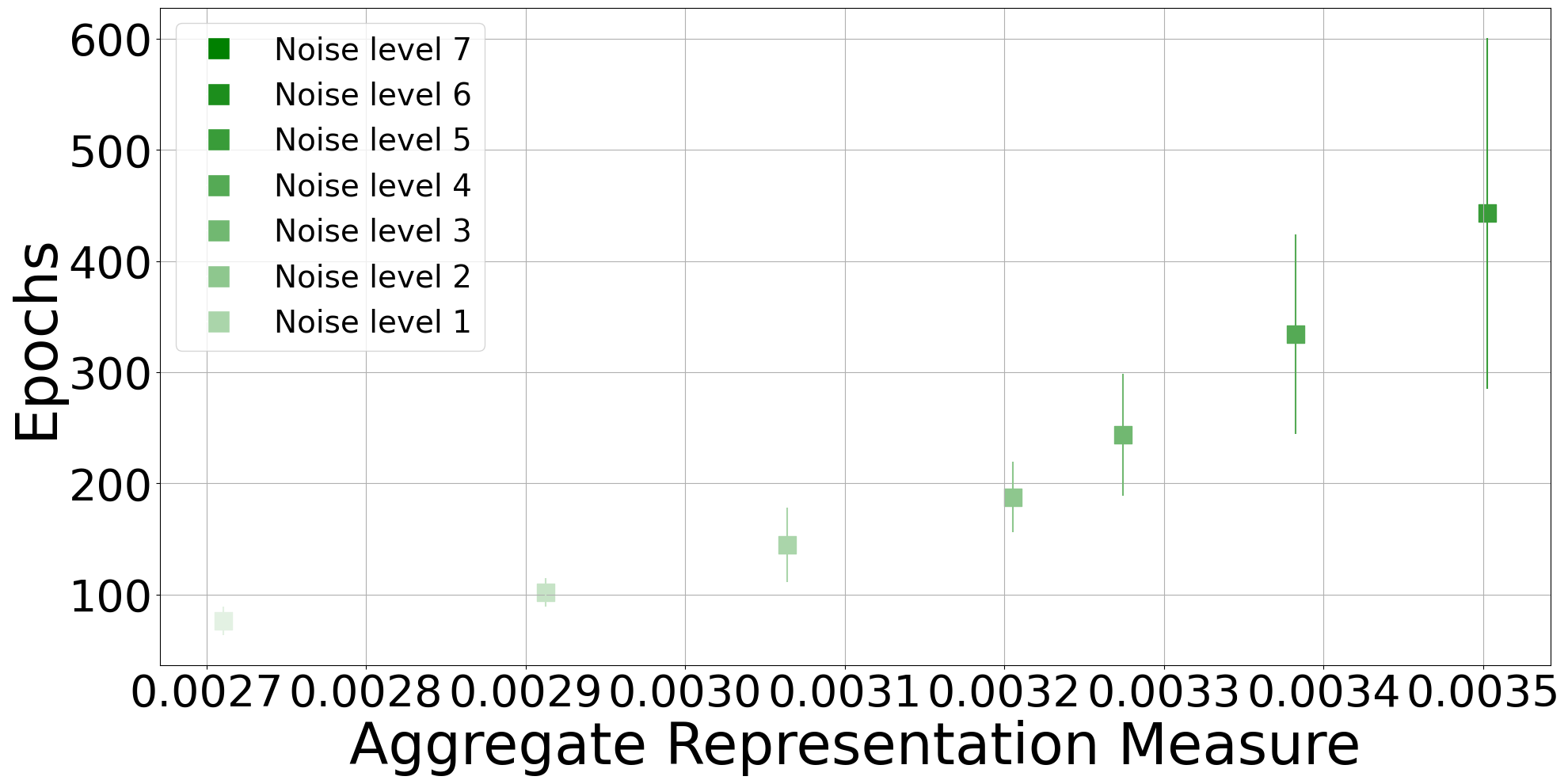}
        \caption{ResNet18 - Gaussian noise}
    \end{subfigure}
    \begin{subfigure}{0.45\textwidth}
        \includegraphics[width=\linewidth]{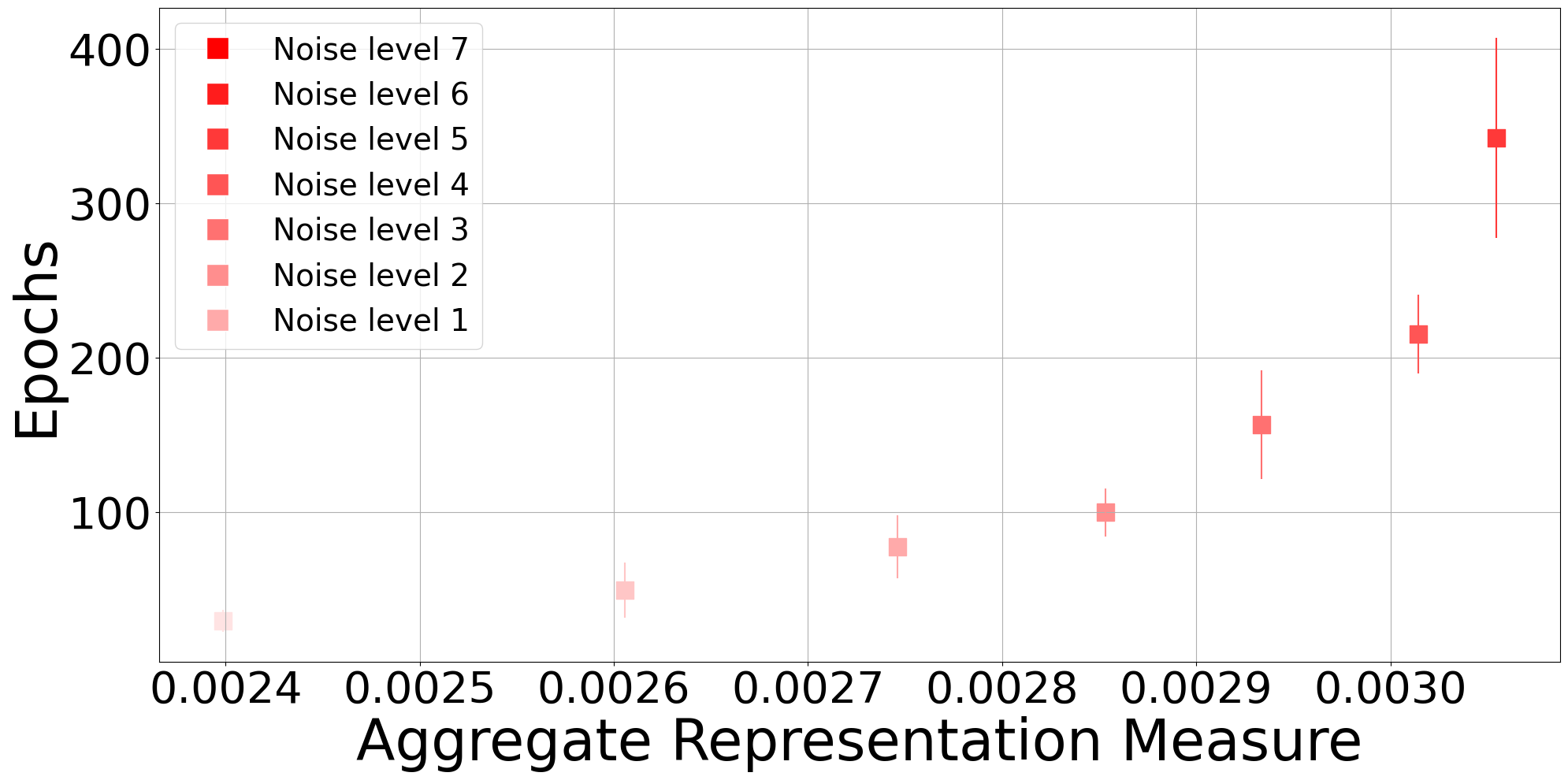}
        \caption{MobileNetv2 - Gaussian noise}
    \end{subfigure}
    \hfill
    \begin{subfigure}{0.45\textwidth}
        \includegraphics[width=\linewidth]{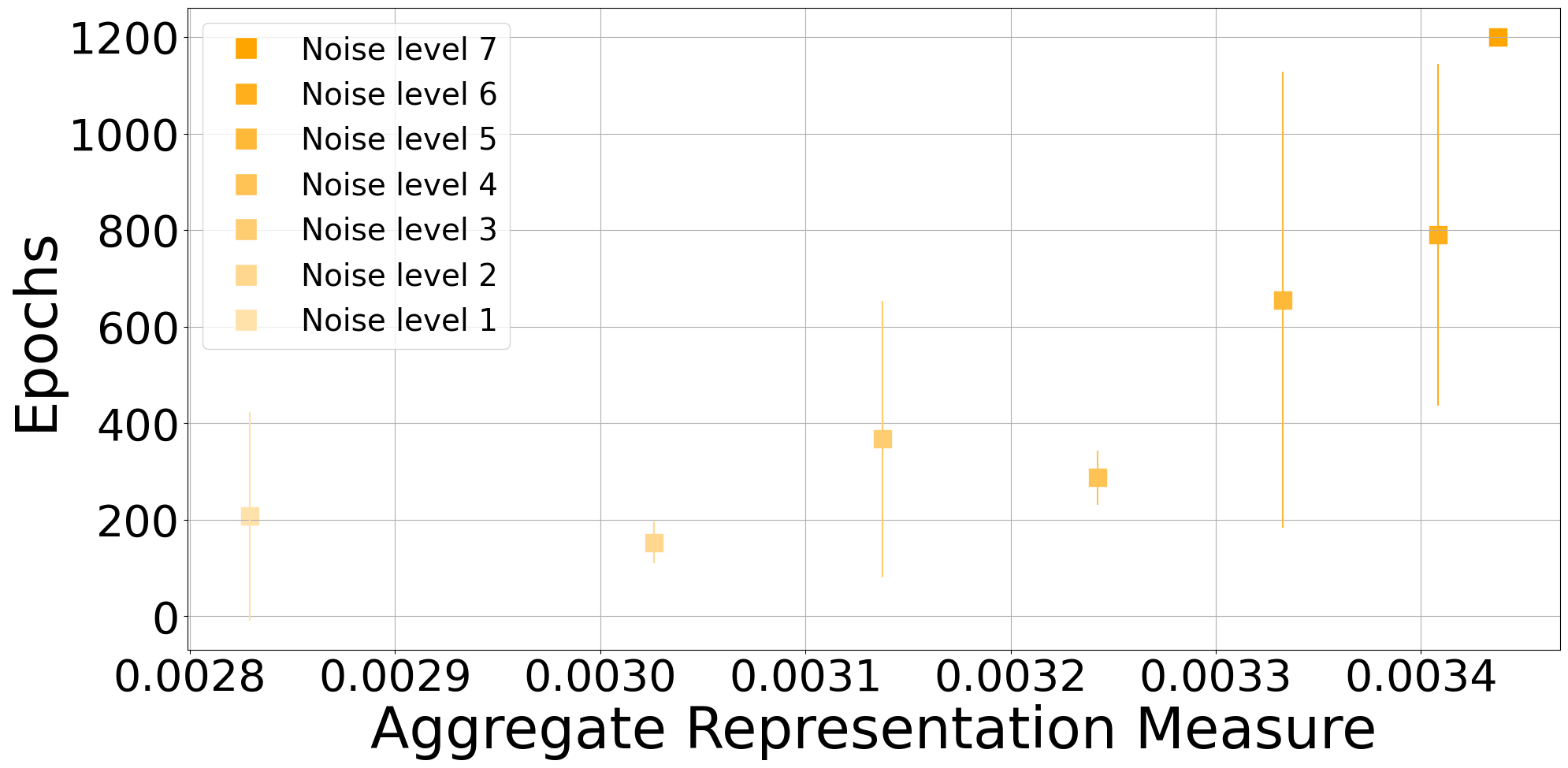}
        \caption{ResNet50 - Gaussian noise}
    \end{subfigure}
    \caption{ARM vs Retraining Epochs on CIFAR100 dataset with Gaussian noise}
    \label{fig:CIFAR100_Gauss}
\end{figure}

\begin{figure}[ht]
    \centering
    \begin{subfigure}{0.45\textwidth}
        \includegraphics[width=\linewidth]{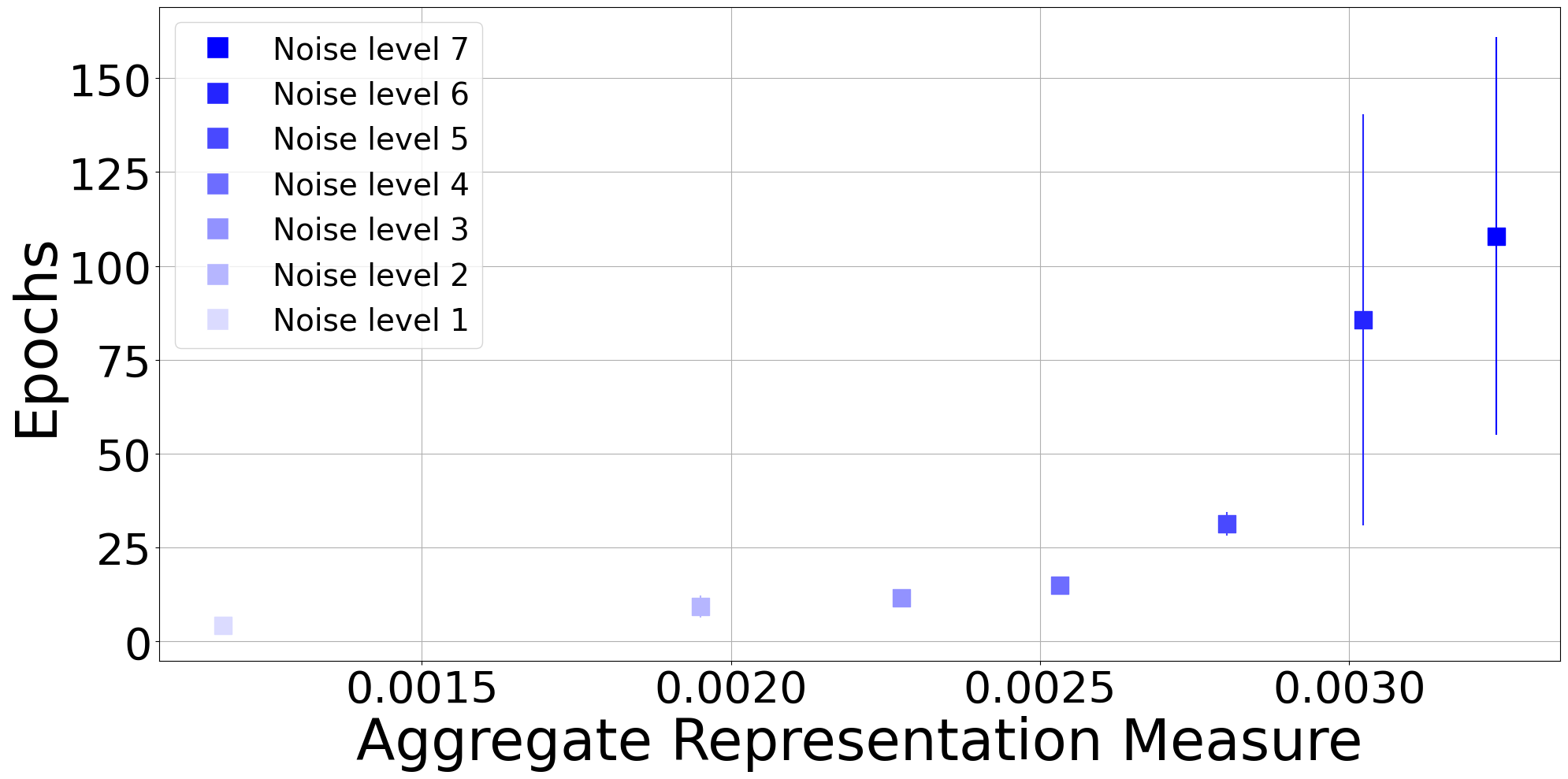}
        \caption{GoogLeNet - Salt-and-Pepper noise}
    \end{subfigure}
    \hfill
    \begin{subfigure}{0.45\textwidth}
        \includegraphics[width=\linewidth]{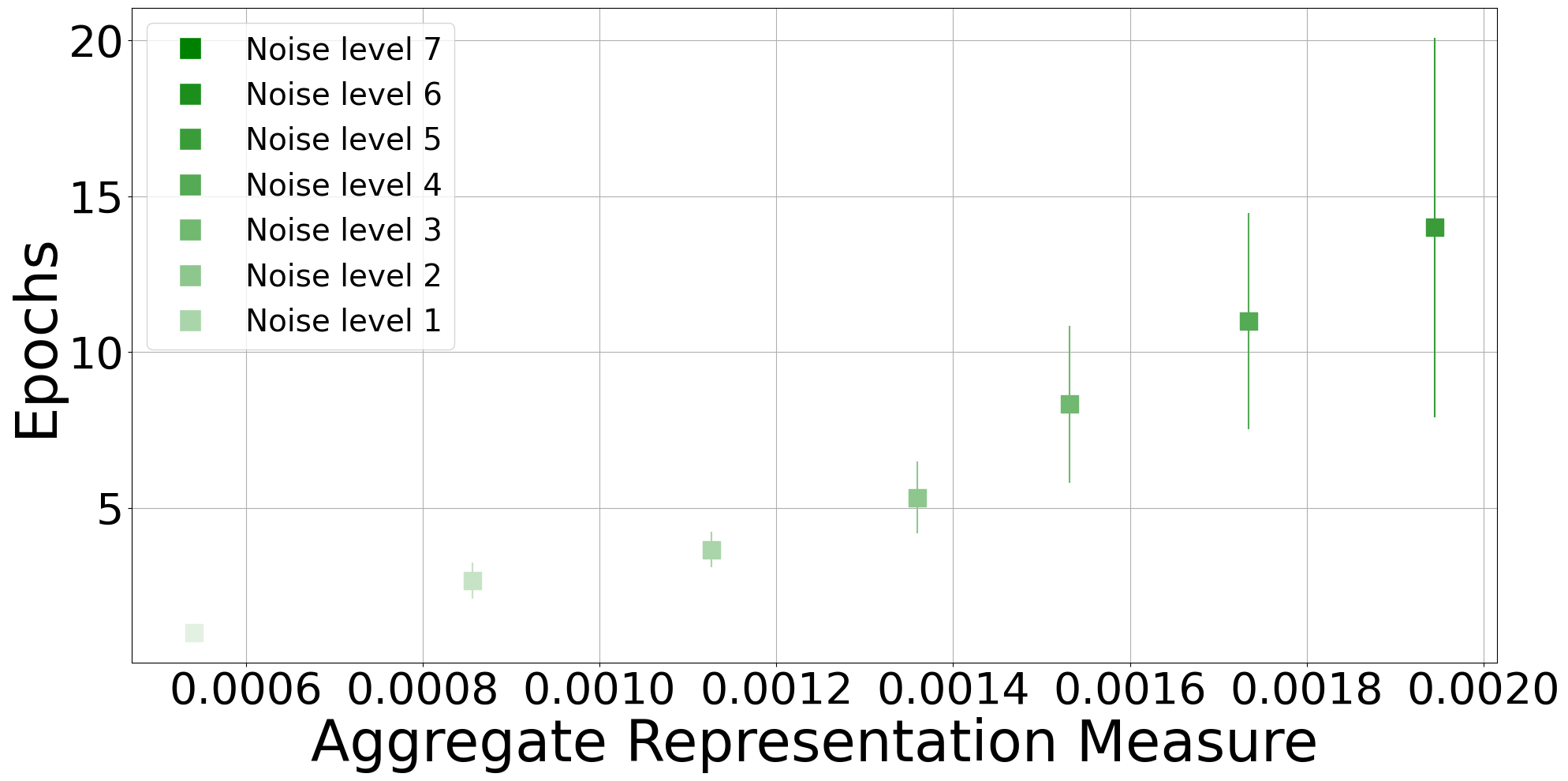}
        \caption{ResNet18 - Salt-and-Pepper noise}
    \end{subfigure}
    \begin{subfigure}{0.45\textwidth}
        \includegraphics[width=\linewidth]{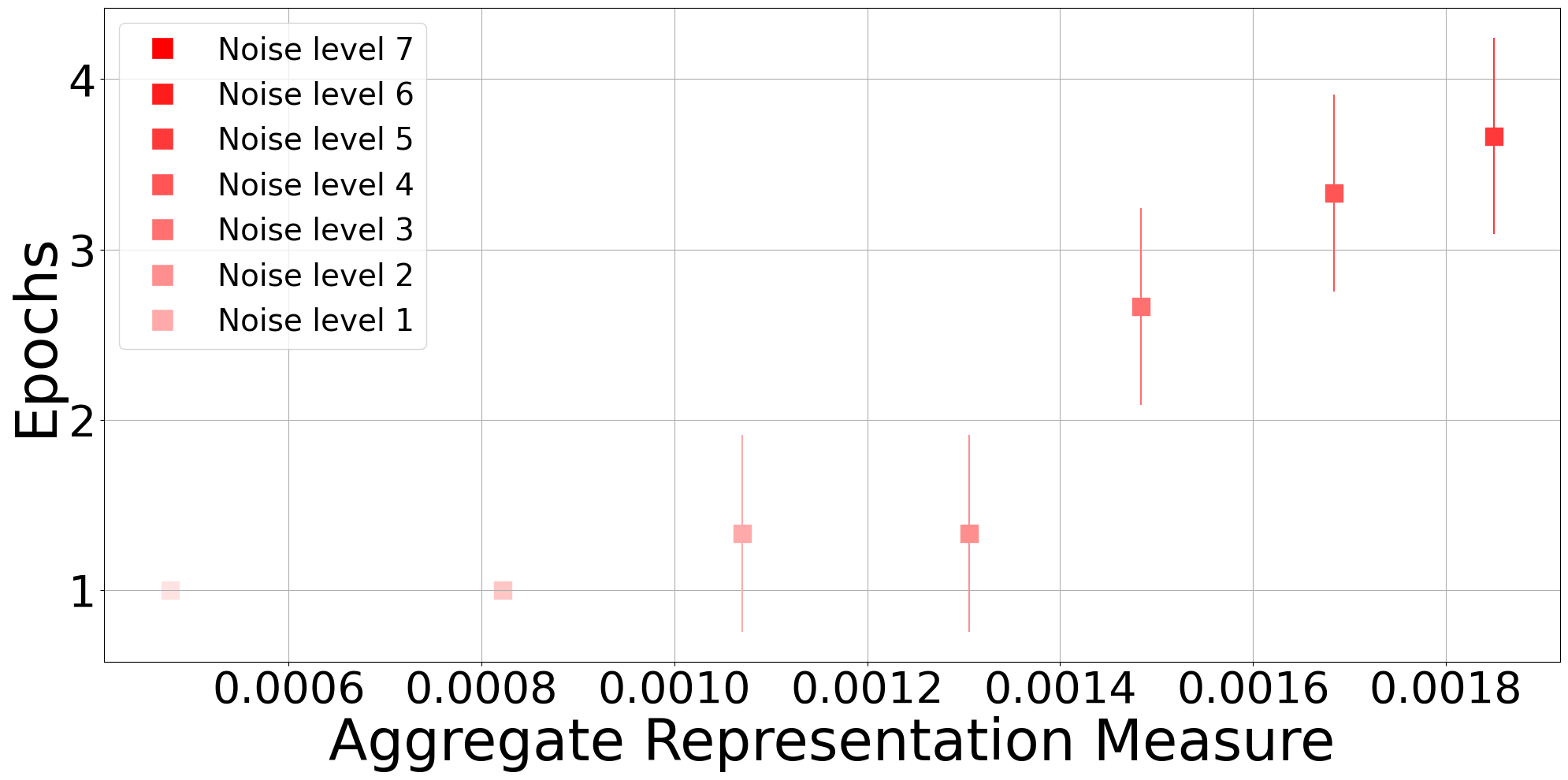}
        \caption{MobileNetv2 - Salt-and-Pepper noise}
    \end{subfigure}
    \hfill
    \begin{subfigure}{0.45\textwidth}
        \includegraphics[width=\linewidth]{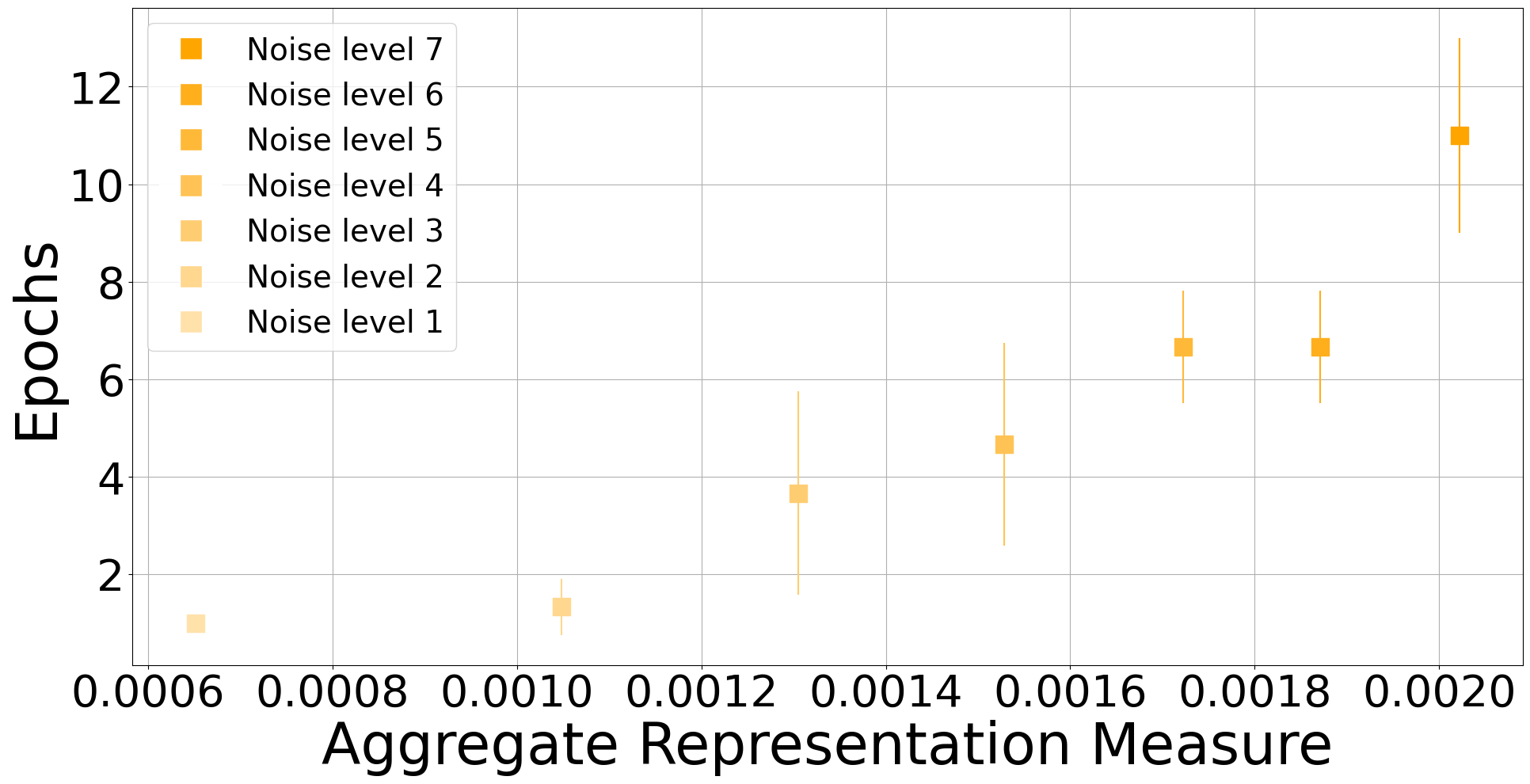}
        \caption{ResNet50 - Salt-and-Pepper noise}
    \end{subfigure}
    \caption{ARM vs Retraining Epochs on CIFAR100 dataset with Salt and Pepper noise}
    \label{fig:CIFAR100_SaltPepper}
\end{figure}

\begin{figure}[ht]
    \centering
    \begin{subfigure}{0.45\textwidth}
        \includegraphics[width=\linewidth]{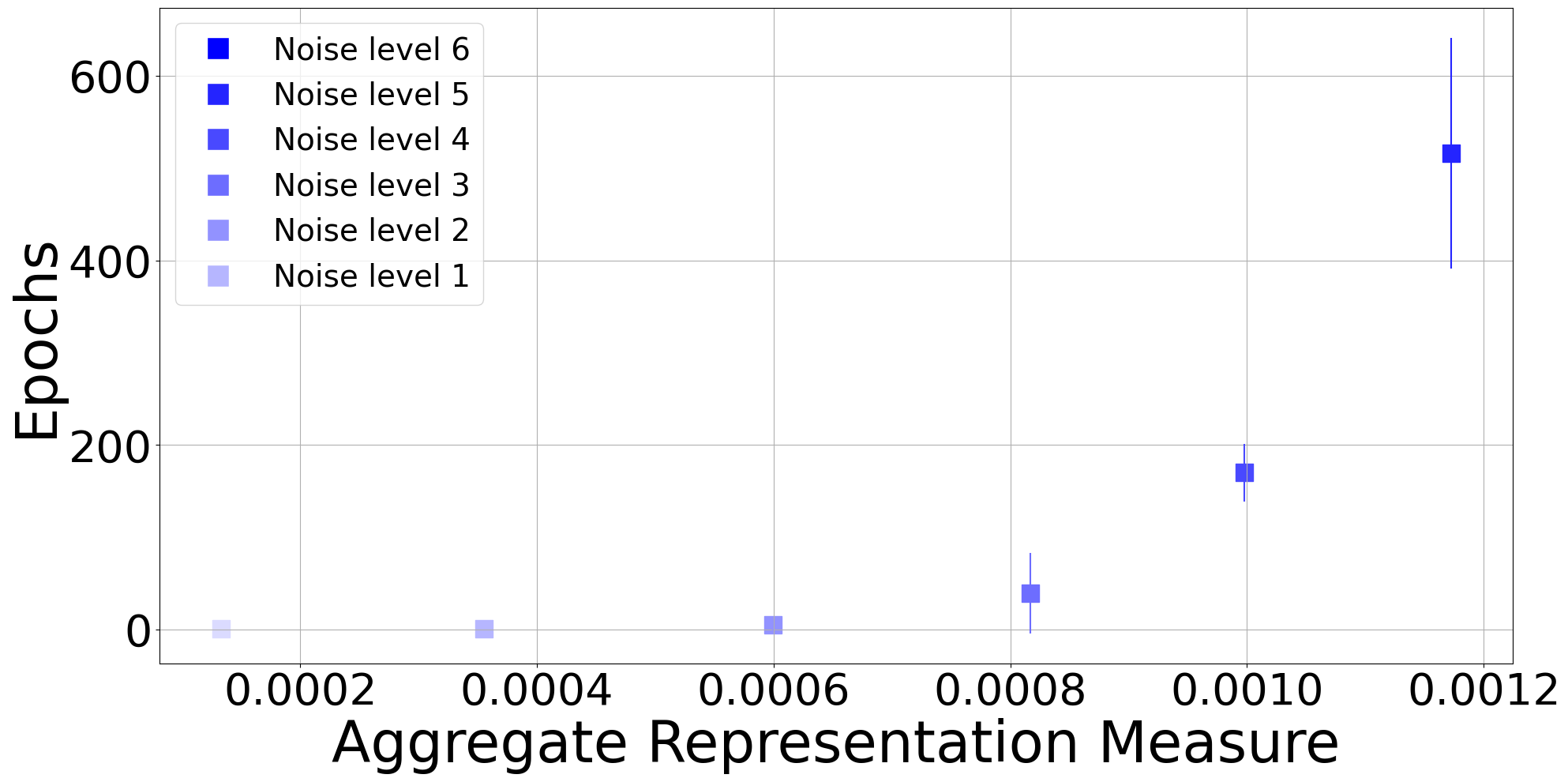}
        \caption{GoogLeNet - Image Blur}
    \end{subfigure}
    \hfill
    \begin{subfigure}{0.45\textwidth}
        \includegraphics[width=\linewidth]{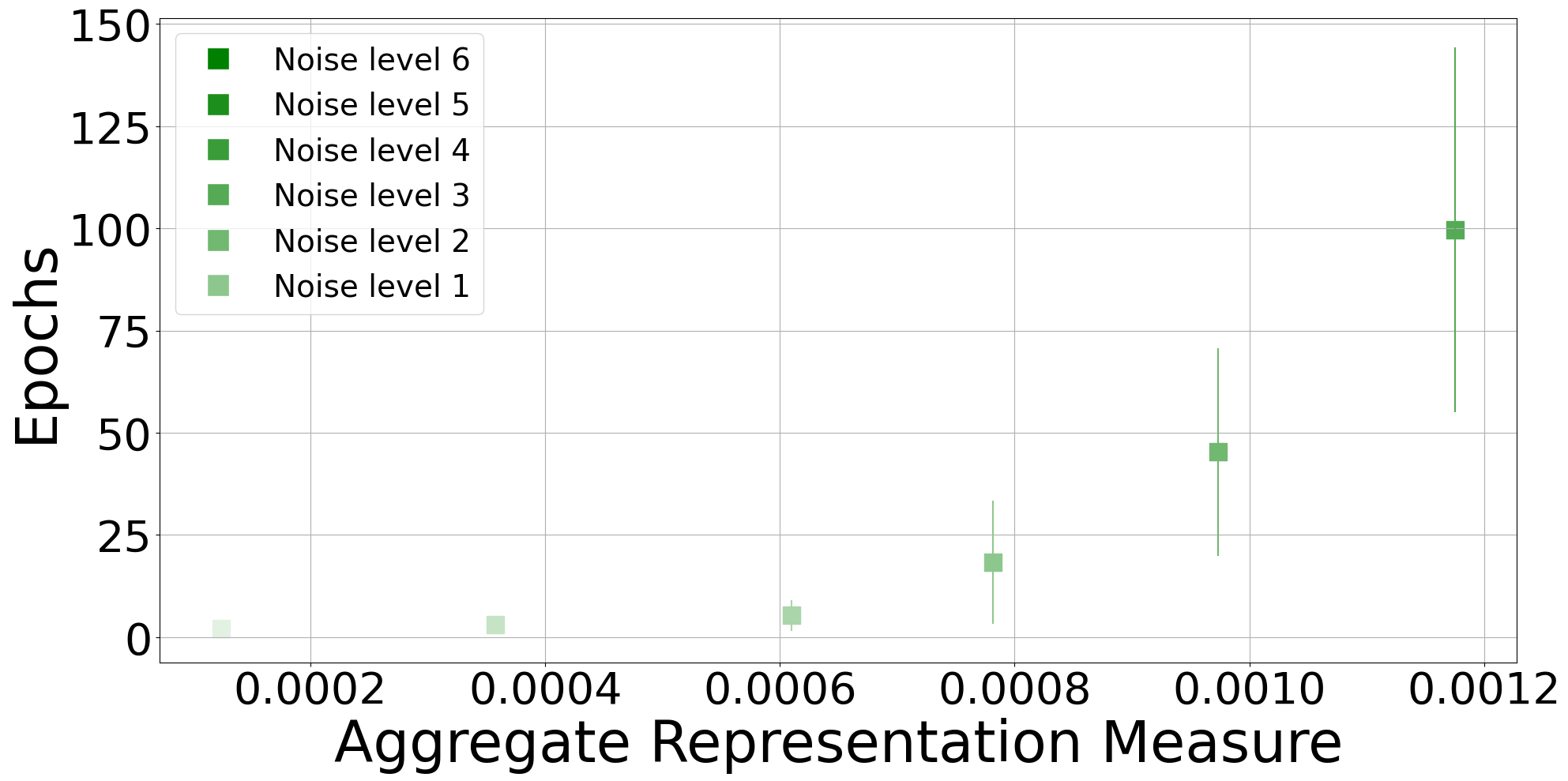}
        \caption{ResNet18 - Image Blur}
    \end{subfigure}
    \begin{subfigure}{0.45\textwidth}
        \includegraphics[width=\linewidth]{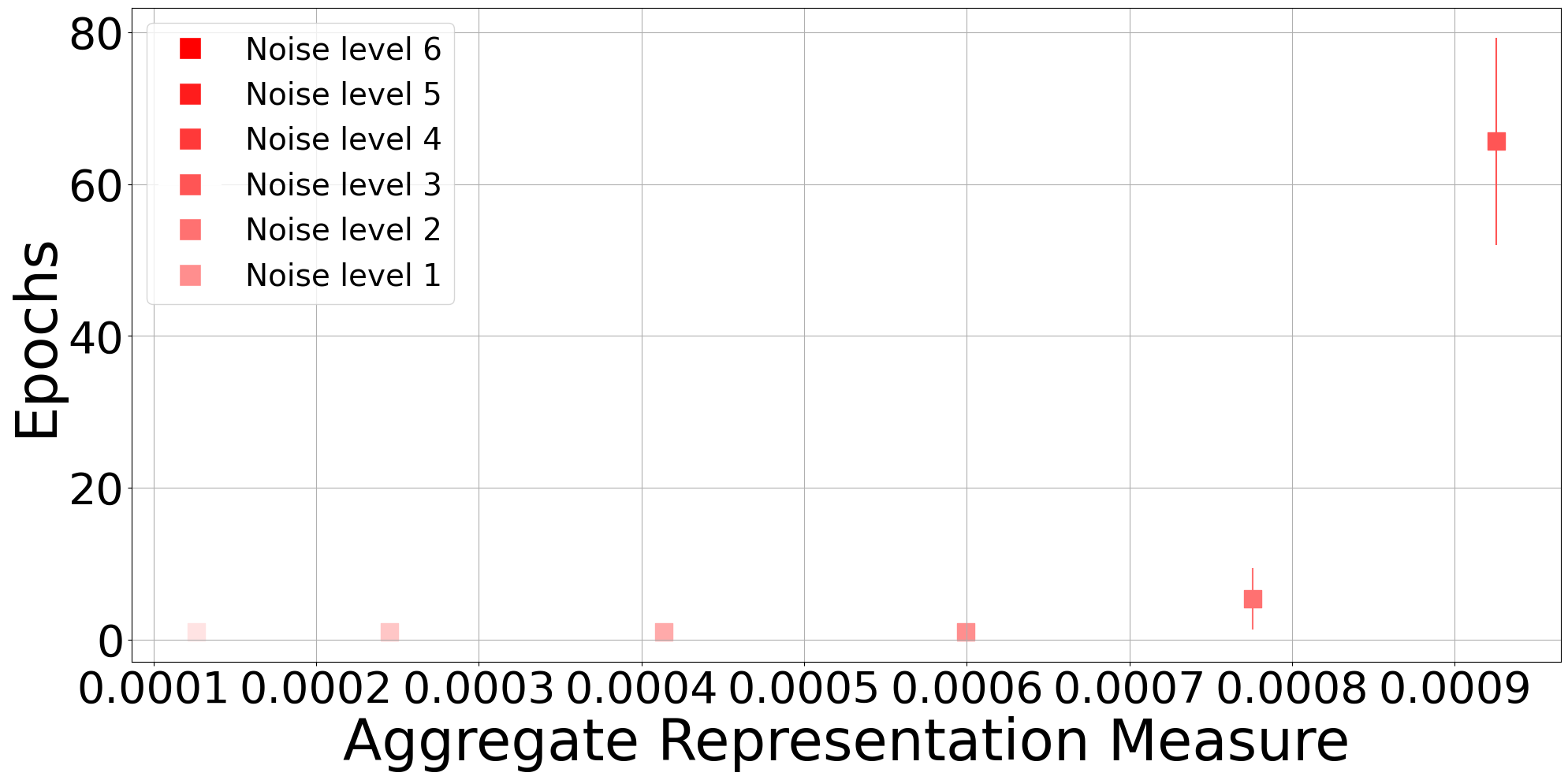}
        \caption{MobileNetv2 - Image Blur}
    \end{subfigure}
    \hfill
    \begin{subfigure}{0.45\textwidth}
        \includegraphics[width=\linewidth]{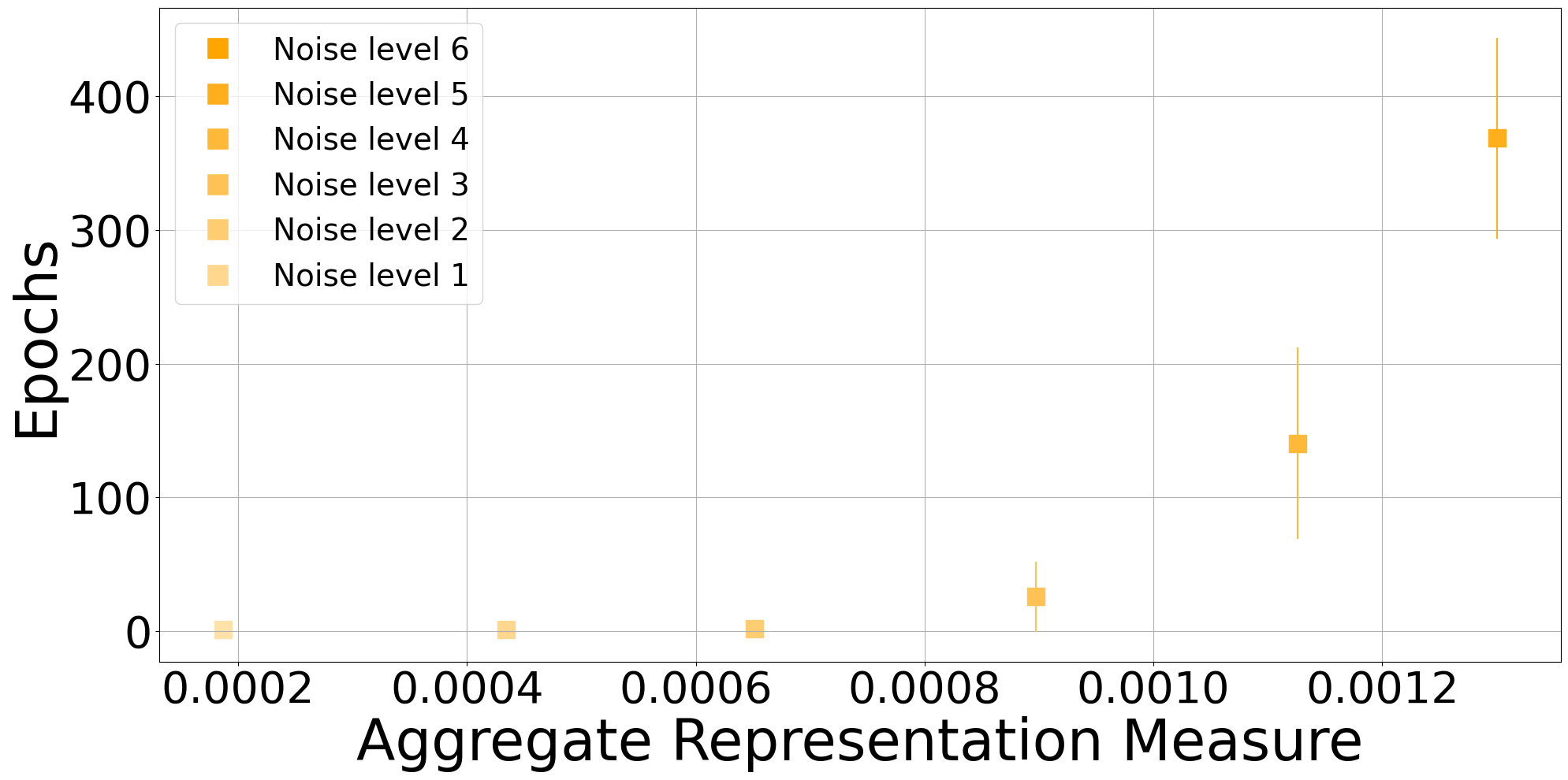}
        \caption{ResNet50 - Image Blur}
    \end{subfigure}
    \caption{ARM vs Retraining Epochs on CIFAR100 dataset with Image Blur}
    \label{fig:CIFAR100_ImageBlur}
\end{figure}

This section provides all results for GoogleNet, ResNet18, MobileNetV2, and ResNet50 on the CIFAR100 dataset. Fig.~\ref{fig:CIFAR100_Gauss}, Fig.~\ref{fig:CIFAR100_SaltPepper} and Fig.~\ref{fig:CIFAR100_ImageBlur} illustrates the ARM and retraining epochs for different levels of Gaussian noise, Salt-and-Pepper noise and Image Blur, respectively. Table.~\ref{tab:correlation_table_C100} provides the Pearson correlation coefficient and p-values for the 4 models retrained on CIFAR100 for the 3 noise types.

\begin{table}[ht]
\small
        \centering
        \begin{tabular}{ccccccc}
             \toprule
             Model & Coefficient & p-value & Coefficient & p-value & Coefficient & p-value \\
             \midrule
             \multicolumn{1}{c}{ } &\multicolumn{2}{c}{Salt-Pepper} & \multicolumn{2}{c}{Gaussian} & \multicolumn{2}{c}{Blur} \\
             \midrule
             GoogLeNet & 0.79 & 0.032 & 0.97 & 0.00023 & 0.78 & 0.066 \\
             \hline
             ResNet18 & 0.96 & 0.0004 & 0.93 & 0.0019 & 0.85 & 0.030 \\
             \hline
             MobileNetV2 & 0.90 & 0.005 & 0.85 & 0.014 & 0.68 & 0.13 \\
             \hline
             ResNet50 & 0.93 & 0.0023 & 0.82 & 0.021 & 0.805 & 0.53 \\
             \bottomrule
        \end{tabular}
        \vspace{0.1cm}
        \caption{Pearson correlation between epochs and ARM - CIFAR100 Dataset}
        \label{tab:correlation_table_C100}
\end{table}

%% file: main.bbl
\begin{thebibliography}{10}

\bibitem{agarwal2022estimating}
Chirag Agarwal, Daniel D'souza, and Sara Hooker.
\newblock Estimating example difficulty using variance of gradients, 2022.

\bibitem{andreassen2021evolution}
Anders Andreassen, Yasaman Bahri, Behnam Neyshabur, and Rebecca Roelofs.
\newblock The evolution of out-of-distribution robustness throughout fine-tuning, 2021.

\bibitem{anthony2020carbontracker}
Lasse F.~Wolff Anthony, Benjamin Kanding, and Raghavendra Selvan.
\newblock Carbontracker: Tracking and predicting the carbon footprint of training deep learning models, 2020.

\bibitem{djolonga2021robustness}
Josip Djolonga, Jessica Yung, Michael Tschannen, Rob Romijnders, Lucas Beyer, Alexander Kolesnikov, Joan Puigcerver, Matthias Minderer, Alexander D'Amour, Dan Moldovan, Sylvain Gelly, Neil Houlsby, Xiaohua Zhai, and Mario Lucic.
\newblock On robustness and transferability of convolutional neural networks, 2021.

\bibitem{drenkow2022systematic}
Nathan Drenkow, Numair Sani, Ilya Shpitser, and Mathias Unberath.
\newblock A systematic review of robustness in deep learning for computer vision: Mind the gap?, 2022.

\bibitem{ford2019adversarial}
Nic Ford, Justin Gilmer, Nicolas Carlini, and Dogus Cubuk.
\newblock Adversarial examples are a natural consequence of test error in noise, 2019.

\bibitem{GARCIAMARTIN201975}
Eva García-Martín, Crefeda~Faviola Rodrigues, Graham Riley, and Håkan Grahn.
\newblock Estimation of energy consumption in machine learning.
\newblock {\em Journal of Parallel and Distributed Computing}, 134:75--88, 2019.

\bibitem{geirhos2020generalisation}
Robert Geirhos, Carlos R.~Medina Temme, Jonas Rauber, Heiko~H. Schütt, Matthias Bethge, and Felix~A. Wichmann.
\newblock Generalisation in humans and deep neural networks, 2020.

\bibitem{gholami2022survey}
Amir Gholami, Sehoon Kim, Zhen Dong, Zhewei Yao, Michael~W Mahoney, and Kurt Keutzer.
\newblock A survey of quantization methods for efficient neural network inference.
\newblock In {\em Low-Power Computer Vision}, pages 291--326. Chapman and Hall/CRC, 2022.

\bibitem{goyal2022testtime}
Sachin Goyal, Mingjie Sun, Aditi Raghunathan, and Zico Kolter.
\newblock Test-time adaptation via conjugate pseudo-labels, 2022.

\bibitem{hendrycks2019benchmarking}
Dan Hendrycks and Thomas Dietterich.
\newblock Benchmarking neural network robustness to common corruptions and perturbations, 2019.

\bibitem{hendrycks2020augmix}
Dan Hendrycks, Norman Mu, Ekin~D. Cubuk, Barret Zoph, Justin Gilmer, and Balaji Lakshminarayanan.
\newblock Augmix: A simple data processing method to improve robustness and uncertainty, 2020.

\bibitem{kimICML20}
Jang-Hyun Kim, Wonho Choo, and Hyun~Oh Song.
\newblock Puzzle mix: Exploiting saliency and local statistics for optimal mixup.
\newblock In {\em International Conference on Machine Learning (ICML)}, 2020.

\bibitem{lacoste2019quantifying}
Alexandre Lacoste, Alexandra Luccioni, Victor Schmidt, and Thomas Dandres.
\newblock Quantifying the carbon emissions of machine learning, 2019.

\bibitem{lee2020smoothmix}
Jin-Ha Lee, Muhammad~Zaigham Zaheer, Marcella Astrid, and Seung-Ik Lee.
\newblock Smoothmix: a simple yet effective data augmentation to train robust classifiers.
\newblock In {\em Proceedings of the IEEE/CVF conference on computer vision and pattern recognition workshops}, pages 756--757, 2020.

\bibitem{lim2023ttn}
Hyesu Lim, Byeonggeun Kim, Jaegul Choo, and Sungha Choi.
\newblock Ttn: A domain-shift aware batch normalization in test-time adaptation, 2023.

\bibitem{liu2022randommix}
Xiaoliang Liu, Furao Shen, Jian Zhao, and Changhai Nie.
\newblock Randommix: A mixed sample data augmentation method with multiple mixed modes, 2022.

\bibitem{niu2022efficient}
Shuaicheng Niu, Jiaxiang Wu, Yifan Zhang, Yaofo Chen, Shijian Zheng, Peilin Zhao, and Mingkui Tan.
\newblock Efficient test-time model adaptation without forgetting, 2022.

\bibitem{schmidt2021codecarbon}
Victor Schmidt, Kamal Goyal, Aditya Joshi, Boris Feld, Liam Conell, Nikolas Laskaris, Doug Blank, Jonathan Wilson, Sorelle Friedler, and Sasha Luccioni.
\newblock Codecarbon: estimate and track carbon emissions from machine learning computing (2021).
\newblock {\em DOI: https://doi. org/10.5281/zenodo}, 4658424, 2021.

\bibitem{stacke2020measuring}
Karin Stacke, Gabriel Eilertsen, Jonas Unger, and Claes Lundstr{\"o}m.
\newblock Measuring domain shift for deep learning in histopathology.
\newblock {\em IEEE journal of biomedical and health informatics}, 25(2):325--336, 2020.

\bibitem{strubell2019energy}
Emma Strubell, Ananya Ganesh, and Andrew McCallum.
\newblock Energy and policy considerations for deep learning in nlp, 2019.

\bibitem{wang2022continual}
Qin Wang, Olga Fink, Luc~Van Gool, and Dengxin Dai.
\newblock Continual test-time domain adaptation, 2022.

\bibitem{xu2021survey}
Jingjing Xu, Wangchunshu Zhou, Zhiyi Fu, Hao Zhou, and Lei Li.
\newblock A survey on green deep learning, 2021.

\bibitem{xu2023energy}
Yinlena Xu, Silverio Martínez-Fernández, Matias Martinez, and Xavier Franch.
\newblock Energy efficiency of training neural network architectures: An empirical study, 2023.

\bibitem{yang2017designing}
Tien-Ju Yang, Yu-Hsin Chen, and Vivienne Sze.
\newblock Designing energy-efficient convolutional neural networks using energy-aware pruning, 2017.

\bibitem{yin2020fourier}
Dong Yin, Raphael~Gontijo Lopes, Jonathon Shlens, Ekin~D. Cubuk, and Justin Gilmer.
\newblock A fourier perspective on model robustness in computer vision, 2020.

\bibitem{zhang2018mixup}
Hongyi Zhang, Moustapha Cisse, Yann~N. Dauphin, and David Lopez-Paz.
\newblock mixup: Beyond empirical risk minimization, 2018.

\end{thebibliography}
